\documentclass{article}

\usepackage{microtype}
\usepackage{graphicx}
\usepackage{subfigure}
\usepackage{booktabs} 
\usepackage[dvipsnames]{xcolor}
\usepackage{hyperref}

\usepackage[accepted]{icml2024}

\usepackage{amssymb}
\usepackage{mathtools}
\usepackage{amsthm}
\usepackage{comment}
\usepackage{multirow}
\usepackage{verbatim}
\usepackage{amsmath}

\usepackage[capitalize,noabbrev]{cleveref}

\usepackage{listings}
\definecolor{codegreen}{rgb}{0,0.6,0}
\definecolor{codegray}{rgb}{0.5,0.5,0.5}
\definecolor{codepurple}{rgb}{0.58,0,0.82}
\definecolor{backcolour}{rgb}{0.95,0.95,0.92}

\lstdefinestyle{mystyle}{
    backgroundcolor=\color{backcolour},   
    commentstyle=\color{codegreen},
    keywordstyle=\color{magenta},
    numberstyle=\tiny\color{codegray},
    stringstyle=\color{codepurple},
    basicstyle=\ttfamily\footnotesize,
    breakatwhitespace=false,         
    breaklines=true,                 
    captionpos=b,                    
    keepspaces=true,                 
    numbers=left,                    
    numbersep=5pt,                  
    showspaces=false,                
    showstringspaces=false,
    showtabs=false,                  
    tabsize=2
}

\lstnewenvironment{promptexample}[1][]
{
    \lstset{style=mystyle, #1}
}
{}

\usepackage{balance}
\usepackage{tikz}
\usetikzlibrary{patterns}
\usepackage{pgfplots}
\pgfplotsset{compat=1.18} 
\usepgflibrary{patterns}
\usepackage{lipsum}

\theoremstyle{plain}

\theoremstyle{definition}

\theoremstyle{remark}

\usepackage[textsize=tiny]{todonotes}

\icmltitlerunning{Automated Evaluation of Retrieval-Augmented Language Models with Task-Specific Exam Generation}

\begin{document}

\twocolumn[
\icmltitle{Automated Evaluation of Retrieval-Augmented \\
            Language Models with Task-Specific Exam Generation}



\begin{icmlauthorlist}
\icmlauthor{Gauthier Guinet}{comp}
\icmlauthor{Behrooz Omidvar-Tehrani}{comp}
\icmlauthor{Anoop Deoras}{comp}
\icmlauthor{Laurent Callot}{comp}
\end{icmlauthorlist}

\icmlcorrespondingauthor{Gauthier Guinet}{guinetgg@amazon.com}

\icmlaffiliation{comp}{AWS AI Labs}

\icmlkeywords{Machine Learning, ICML, Large Language Models, Evaluation}

\vskip 0.3in]


\printAffiliationsAndNotice{}  

\begin{abstract}
We propose a new method to measure the task-specific accuracy of Retrieval-Augmented Large Language Models (RAG). Evaluation is performed by scoring the RAG on an automatically-generated synthetic exam composed of multiple choice questions based on the corpus of documents associated with the task. Our method is an automated, cost-efficient, interpretable, and robust strategy to select the optimal components for a RAG system. 
We leverage Item Response Theory (IRT) to estimate the quality of an exam and its informativeness on task-specific accuracy. IRT also provides a natural way to iteratively improve the exam by eliminating the exam questions that are not sufficiently informative about a model's ability. 
We demonstrate our approach on four new open-ended Question-Answering tasks based on Arxiv abstracts, StackExchange questions, AWS DevOps troubleshooting guides, and SEC filings. In addition, our experiments reveal more general insights into factors impacting RAG performance like size, retrieval mechanism, prompting and fine-tuning. Most notably, our findings show that choosing the right retrieval algorithms often leads to bigger performance gains than simply using a larger language model.

\end{abstract}

\section{Introduction}
\label{sec:intro}

\begin{figure}[t]
    \centering
    \includegraphics[width=\linewidth]{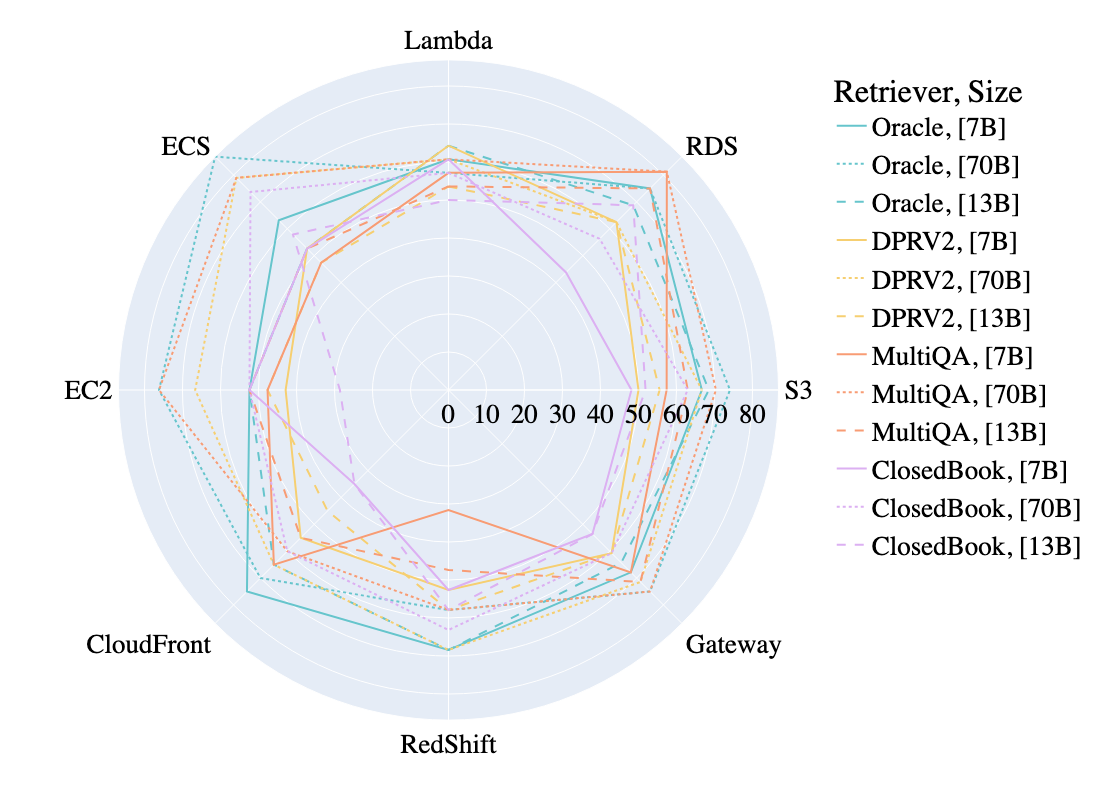}
    \caption{Granular results of our exam evaluation for the task of AWS DevOps troubleshooting. Accuracy is reported for different retrieval approaches and retriever sizes, on a~\% scale. Labels on the diameter shows the troubleshooting categories, i.e., AWS resources. Colors correspond to different retrieval approaches (\texttt{Oracle}, \texttt{DPRV2}, \texttt{MultiQA}, \texttt{ClosedBook}, as discussed in Section~\ref{sec:ragpipelines}) and patterns correspond to the base LLM size (7B, 13B, and 70B). For instance, we observe that a small model such as Mistral-7B with MultiQA embeddings has an accuracy around 80\% for the AWS resource ``Relational Database Service'' (RDS).}
    \label{fig:devOps}
    \vspace{-10pt}
\end{figure}

Evaluating Large Language Models (LLM) beyond a set of limited tasks is notoriously challenging. General capabilities of LLMs conveyed through public benchmarks are not necessarily related to performance on narrow and highly specific customer tasks, even more so when such tasks involve specific domain knowledge corpus. Evaluation metrics aim at capturing different aspects of the performance of a LLM. No single metric can adequately capture every aspects of it.

In this work, we propose an LLM-driven exam-based evaluation methodology to measure the accuracy of Retrieval-Augmented LLMs (RAG) on a given task. Our method is fully automated and does not require annotated ground-truth dataset. Our metrics focus on factual accuracy, i.e., the ability to retrieve and leverage the right information to correctly answer a user query. 
In addition to enabling users to select the optimal combination of components of a RAG system for their retrieval tasks, our methodology reveals more general insights into RAG performance factors like size, retrieval mechanism, prompting, and fine-tuning.

Our first contribution is a quantitative exam-based evaluation strategy which is fully automated, guaranteeing scalability in contrast to conventional human-in-the-loop evaluations, while concurrently mitigating expenses linked with expert or annotator engagement. Exams are generated by a LLM using the corpus of the data associated with the task at hand. Candidate RAG systems are evaluated on their ability to successfully answer the multiple-choice questions in the exam. Evaluation is always a trade-off between ease of scoring and representativeness.  For factual evaluation, the ease of scoring multiple-choice exams does not compromise the objective of assessing factual knowledge efficiently and reliably. Moreover, comparative exam result analysis reveals areas for performance improvement, enabling continuous feedback-driven enhancements to the exam corpus. Figure~\ref{fig:devOps} shows an example in the DevOps domain.

Our second contribution is a methodological improvement strategy within the automated exam generation process. Notably, we draw on Item Response Theory (IRT) to optimize the generated exam and maximize its informativeness on task-specific model performance.

We illustrate and evaluate our approach on open-ended question-answering tasks using 4 different knowledge corpora: AWS DevOps troubleshooting guides, Arxiv abstracts, StackExchange questions, and SEC Filings. In summary, here is the list of our contributions.

\begin{itemize}
    \item We contribute a comprehensive methodology for \textit{automatic evaluation} of Retrieval-Augmented Generation LLM pipelines based on task-specific synthetic exams.
    \item Leveraging Item Response Theory (IRT), we develop \textit{robustness} and \textit{interpretable} evaluation metrics to quantify and elucidate factors influencing model efficacy.  
    \item We design a principled, fully automated technique to \textit{construct} and \textit{iteratively refine} the exams to maximize informativeness. 
    \item We provide benchmark datasets for RAG systems evaluation, by creating four new tasks based on public datasets from diverse domains. 
    \item We provide an open-source implementation of our proposed exam generation, evaluation and optimization framework allowing it to be executed on any RAG task. The source code is available at \href{https://github.com/amazon-science/auto-rag-eval}{https://github.com/amazon-science/auto-rag-eval}.
\end{itemize}

The paper is organized as follows: we review related work in Section~\ref{sec:lit}. In Section~\ref{sec:model}, we discuss the problem of evaluating RAG pipelines, and propose two evaluation modalities. We introduce an extensive benchmark in Section~\ref{sec:benchmark} and present our experiments in Sections~\ref{sec:exp} (model evaluation) and~\ref{sec:examresults} (exam evaluation). We conclude in Section~\ref{sec:conc}.
\section{Related Literature}
\label{sec:lit}
We propose an automated exam generation method that enables standardized evaluation of RAG on specific tasks by tailoring multiple choice questions to each task's documents. To the best of our knowledge, this is the first work addressing RAG assessment with this particular contribution focus. However, our contribution builds upon existing literature in related domains like RAG systems, evaluation frameworks for language models, and item response theory.

\noindent \textbf{Retrieval-Augmented Generation.} 
Retrieval-Augmented Generation (RAG) integrates pre-trained language models with information retrieval techniques, enriching natural language processing tasks through external knowledge sources~\citep{DBLP:journals/corr/abs-2005-11401}. This methodology was further developed by~\citep{DBLP:journals/corr/abs-1911-00172} who highlighted the effectiveness of nearest-neighbor search within language models. Subsequent advancements include the introduction of a self-supervised learning objective, synergizing language models with retrieval systems~\citep{DBLP:journals/corr/abs-2002-08909}, and the expansion of retrieval-augmented methods for handling larger data scales~\citep{DBLP:journals/corr/abs-2112-04426}. The field is comprehensively surveyed by \citep{gao2023retrieval}. 

\noindent \textbf{Evaluation of NLP, LLM, and RAG.} The evolution of evaluation in Natural Language Processing (NLP) has transitioned from classical, task-specific benchmarks, like BLEU scores for machine translation~\citep{papineni2002machine}, to more nuanced metrics~\cite{es2023ragas,zheng2023judging,hoshi2023ralle,saad2023ares}, like Answer Equivalence~\cite{bulian2022tomayto}, as the complexity of outputs has increased. This shift is exemplified by the work on GPT-3 which challenges traditional evaluation methods with its task-agnostic capabilities~\citep{brown2020language, bender2020climbing}. Notwithstanding these advances, the field continues to confront well documented challenges \citep{deutsch2021statistical, bowman2021will, bulian2022tomayto, novikova2017we, fabbri2021summeval}, in particular to accurately measure models' understanding of nuanced human concepts.

When evaluating retrieval-augmented generation models, the difficulties are compounded by the multiplicity of components involved. \citep{gao2023retrieval} offers a survey of the field, revealing \textit{``a notable paucity of research dedicated to evaluating the distinct characteristics of RAG models, with only a handful of related studies.''} Most recent work in this domain emphasize the integration of retrieved information with generated content~\citep{DBLP:journals/corr/abs-2005-11401, kamalloo2023hagrid, chen2023benchmarking}. Some solutions like RAGAs \citep{es2023ragas}, RaLLe \citep{hoshi2023ralle}, and ARES \citep{saad2023ares} are being increasingly used in industrial and research applications but offer limited interpretability. Furthermore, although certain benchmarks have been designed to assess specific aspects of LLMs (e.g, truthfulness~\cite{lin2021truthfulqa}, faithfulness~\cite{adlakha2023evaluating}, and factuality~\cite{lee2022factuality,min2023factscore,yu2023kola,muhlgay2023generating}), a comprehensive task-specific evaluation of these aspects remains a significant challenge. The absence of a canonical evaluation method for RAG pipelines is the main driver of our contribution in this work.


\noindent \textbf{Item Response Theory.} Item Response Theory (IRT) is a framework to study responses to questions (items) in evaluations conducted through examination. IRT offers a model-based approach for estimating both item characteristics and individual examinee abilities. Its formalization appears to originate with the works of \cite{rasch1960studies, lord1968statistical}. This theory has been developed and extended to be applied to a wide range of problems which are reviewed in \cite{embretson2013item, cai2016item} among others. 
In machine learning, IRT has found applications for providing interpretability \citep{yeung2019deep, martinez2019item, martinez2016making}, improving recommender systems \cite{liu2023we}, or guiding human evaluation of chatbots \cite{sedoc2020item}.
To the best of our knowledge, this paper is the first one to leverage item response theory in order to develop an automated evaluation procedure for generative models.
\section{Methodology}
\label{sec:model}
In this section, we define the key concepts upon which we build our contributions, discuss the problem of evaluation of RAG pipelines, and propose two evaluation modalities.  

\subsection{Preliminaries}

\noindent \textbf{RAG pipelines.} We consider a RAG pipeline to be constituted by three components: the LLM, the retrieval mechanism, and the in-context learning part. 
First is the LLM which is used to generate an answer given some retrieved context and a prompting strategy. We rely on broadly available, pre-trained large language models.
The second component is a retrieval mechanism which is used to identify documents in the corpus relevant to the user's question. These documents are then included in the LLM's prompt to provide helpful context for answering.
Finally, the third component is the in-context learning part of the prompt given to the LLM. In this paper, the in-context learning mechanism is the number of examples of the task we provide in the prompt. Note that we could incorporate more complex RAG design choices: data processing, query refactoring, more elaborated prompting, fine-tuning, and post-generation processing. However, in the sake of generality, we focus on the three aforementioned choices and remark that our approach easily extends to other settings.

\noindent \textbf{Tasks.} The generic task we consider here is that of open-ended question-answering supported by a corpus of documents in which the answer is expected to be found. A task $t \in \mathcal{T}$ is characterized by a knowledge corpus which is composed of set of documents from a specific domain. The retrieval mechanism extracts from the corpus the documents that are most relevant to answering the user's question.

\noindent \textbf{Evaluation.} Evaluation should be seen through two lenses: \textit{predictive} and \textit{prescriptive}. The goal of a predictive evaluation is to design an estimator of the accuracy on a downstream task of interest. Prescriptive evaluation guides design decision by providing insights on the choice of model to make, as well as the impact of the different components. Our main contribution in this work, the exam-based evaluation methodology, is used for both predictive and prescriptive evaluation.  
For predictive evaluation, each RAG pipeline is evaluated independent from other pipelines by answering an exam composed of multiple-choice questions. This evaluation metric does not quantify all possible dimensions of interest, no single metric does. Our method is predictive in what is arguably the most important performance dimension for a RAG pipeline: the ability to retrieve and leverage external information.
Prescriptive evaluation involves jointly examining multiple pipelines to understand broader patterns. This allows for model ranking and selection and reveals general insights on the drivers of RAG pipeline performance to guide design decisions.

\subsection{Exam Generation}
\label{sec:exam-gen}

The exam generator algorithm leverages a pre-trained LLM which generates a multi-choice exam with $n$ questions for a given task $t$. The output $Q = \{q_1, q_2 \dots q_n\}$ is a set of questions. Each question is composed of a question description and a set of possible answers. There is one and only one correct answer among the possible answers. We leverage here a two-step approach: for each document in the knowledge corpus, we use the LLM and several prompt strategies to create candidates questions. This raw generation is insufficient to generate a high quality exam and thus we combined it with several NLP-based filters to remove low-quality questions along several axis such as length, incorrectness, and self-containment. We refer to this improvement steps as \textit{a-priori verification} as the filters do not require candidate model answers. In particular, we note an interesting asymmetry: granted a document corpus, it is relatively easy for a LLM to generate a question and the correct answer, as this task is self-contained in the prompt in terms of knowledge. However, it is considerably more difficult to create high quality incorrect answers, commonly referred as discriminators. We leverage Jaccard and embedding based similarity metrics to filter out degenerated questions following this pattern. This methodology and the exam generation process is further detailed in Appendix~\ref{sec:exam-details}.

Throughout the work, we aimed at balancing each contributions on the automated generation of the exam, with equal new methodological contributions to both assess the quality and the evaluation impact of the exam. Among other, the next section introduces a novel \textit{a-posteriori verification}, using Item Response Theory (IRT) to weight the contribution of each question towards the final model score by the inferred quality of the question. This ensures that our evaluation methodology is more robust to outliers and low quality questions.

\subsection{Exam Evaluation}
\label{sec:exam-eval}

\noindent \textbf{Pointwise Evaluation.} To evaluate the performance of a RAG pipeline, we first treat it as a student who participates in an exam generated as described above, where we select for each question the answer with maximal length-penalized log-likelihood \citep{eval-harness}. The score obtained by the RAG is simply the share of questions answered correctly. This examination modality allows us to order RAG pipelines as a function of their performance on a given exam generated using the corpus of documents associated with task $t$. 

This simple exam-based evaluation modality is an automated, scalable, and computationally efficient way to obtain a performance ranking of RAG pipelines designed for a specific task. The performance metric, the grade of a given RAG, is trivially interpretable. Our approach to evaluation is capable of delivering deeper insights, which we discuss next.

\noindent \textbf{Aggregate Evaluation and Item Response Theory.} The aggregate evaluation method \textit{jointly and simultaneously} evaluates multiple RAG pipelines together with the \textit{quality} of the exam $Q$ generated for task $t$ on which the RAGs are graded. This allows to $(i)$ increase robustness by providing weighted RAG ability scores that account for noisy or uninformative questions, $(ii)$ reliably quantify the contribution of each individual RAG components on the final performance and $(iii)$ quantify the exam informativeness over the task of interest. This last point is core in providing a set of quantitative exam analytics with high interpretability (Section~\ref{sec:exam-informativeness} and ~\ref{sec:cat-exam-qna}) and iteratively improve the exam to maximize informativeness (Section~\ref{sec:iterative-improvement}).

To do so, we rely on Item Response Theory (IRT), a modern framework used to understand how exam-takers interact with individual items (i.e., questions) in an exam. 
Item Response Theory models the probability of a correct answer to an exam item $q_i\in Q$ as a function of the exam-taker's ability $\theta$ and of three parameters characterizing a specific question $q_{i}$: difficulty $b_i$, discrimination $d_i$, and guessing factor $g_i$, thanks to the logistic model:
\begin{equation}
\label{eq:irt}
    \mathbb{P}(X = 1 | \theta, g_i, d_i, b_i) = g_i + \frac{(1-g_i)}{1+\mathit{exp}(-d_i(\theta-b_i)))},
\end{equation}
where $X = \{1, 0\}$ indicates a correct or incorrect answer. On what follows, we use the abbreviation $p_{i}(\theta)$ for this quantity, omitting the dependency on $g_i, d_i, b_i$.

The capability of a question to distinguish between student of a given ability $\theta$ is captured by the difficulty parameter. Intuitively, an easy question (low $d_i$) will be answered correctly by all high-ability (high $\theta$) students so it does not help distinguishing the best among those. A question with a high discrimination value $d_i$ amplifies the difference in ability, meaning that the question is better at distinguishing between students that have close but different ability. In all multiple choice questions, there exists a probability to answer the question correctly by chance which is captured by $g_i$.

In this paper, we propose a variation of the standard IRT model of Equation \ref{eq:irt} tailored to the task of evaluating RAG systems, which we call the \textit{hierarchical} IRT model.
The hierarchical model provides a higher resolution estimate of the ability of the RAG by breaking it down into its three components using the additive model $\theta_{m} = \theta_{\mathit{llm}(m)} + \theta_{\mathit{ret}(m)} + \theta_{\mathit{icl}(m)}$. The three parameters quantify the ability of the LLM, retrieval method, and in-context learning method, respectively. Extending this model to more complex RAG design choices only requires adding suited latent variables.

The hierarchical IRT model is one of the key contributions of this paper. It allows us to evaluate the performance of the components of a RAG pipeline independently, which simplifies the problem of model selection substantially. In addition it allows us to derive some general insights on the main drivers of the performance of RAG pipelines, discussed in details in Section \ref{sec:exp}.

\subsection{Item Response Model Estimation}
\label{sec:irt-estimation}

To fit the IRT model, we employ a log-likelihood optimization model to estimate the ability $\theta_{m}$ of the candidate models $m\in\mathcal{M}$, and to jointly estimate the three parameters $\{g_i, d_i, b_i\}$ characterizing each question $q_{i}\in\mathcal{Q}$ in the exam corpus. We maximize the log-likelihood function $\mathcal{L}$ over the parameters $\{\theta_m\}_{m\in\mathcal{M}}$ and $\{g_i, d_i, b_i\}_{q_i\in\mathcal{Q}}$ using the probability function $p_{i}(\theta)$ defined in Equation~\ref{eq:irt}.
\begin{equation}
\label{eq:log-irt}
    \mathcal{L} = \sum_{\substack{m\in\mathcal{M} \\ q_{i}\in\mathcal{Q}}}
 r_{i, m}\log p_{i}(\theta) + (1-r_{i, m})\log(1-p_{i}(\theta)),
\end{equation}

In Equation~\ref{eq:log-irt}, $r_{i,m}$ is a binary function indicating whether model $m$ provided the correct response to question $i$ ($r_{i,m}=1$) or incorrect ($r_{i,m}=0$). For the hierarchical IRT model, we decompose $\theta_{m}$ as $\theta_{\mathit{llm}(m)} + \theta_{\mathit{ret}(m)} + \theta_{\mathit{icl}(m)}$ and maximize over this new space of latent variables. We further detail the estimation procedure and results in Appendix~\ref{sec:hierarchical-irt}. A model is considered to possess high ability if it can accurately respond to challenging questions. Conversely, difficulty questions are deemed so if only students with a high level of competence can answer them. This interdependent problem is what is addressed when maximising Equation~\ref{eq:log-irt}.

\section{Experiment Benchmark}
\label{sec:benchmark}
In this section, we introduce an extensive benchmark instantiated based on the model defined in Section~\ref{sec:model}. We report the experiment results over this benchmark in Section~\ref{sec:exp}.

\begin{table*}[t]
\centering
\begin{tabular}{lrrrr}
\toprule
\textbf{Attribute} & $t_{\mathit{ops}}$ (\textbf{DevOps}) & $t_{\mathit{arx}}$ (\textbf{Arxiv}) & $t_{\mathit{stk}}$ (\textbf{StackExchange}) & $t_{\mathit{sec}}$ (\textbf{SEC Filings}) \\
\midrule
\# Documents & $1249$ webpages & $13\,000$ abstracts & $977$ Stack questions & $493$ sections  \\
\# Documents Chunks & $4\,536$ & $13\,000$ & $977$ & $11\,658$ \\
Average Document Length (words) & $254$ & $189$ & $144$ & $187$ \\
Total \# Words & $1\,153\,149$ & $2\,459\,804$ & $140\,859$ & $2\,175\,250$\\
Vocabulary Size & $9\,175$ & $39\,551$ & $44\,084$ & $11\,229$\\
\# Topics & $18$ & $13$ & $20$ & $10$\\
\bottomrule
\end{tabular}
\caption{Description of all four tasks used in the experiment benchmark. Word count is computed using NLKT word tokenizer and punctuation remover \cite{bird2009natural}.}
\label{tbl:tasks}
\end{table*}

\subsection{Tasks}
\label{sec:tasks}
We introduce four different tasks in our benchmark $\mathcal{T} = \{t_{\mathit{ops}}, t_{\mathit{arx}}, t_{\mathit{stk}}, t_{\mathit{sec}}\}$. The task~$t_{\mathit{ops}}$ is defined over a knowledge corpus of 1249 webpages from AWS Knowledge Center\footnote{\it AWS Knowledge Center: https://repost.aws/knowledge-center.} where each webpage troubleshoots one DevOps problem for AWS customers. The task~$t_{\mathit{arx}}$ is defined over 13000 ArXiv papers where each paper is represented by its abstract. The task~$t_{\mathit{stk}}$ is defined over 977 StackExchange\footnote{\it StackExchange network: https://stackexchange.com/.} questions. Last, the task $t_{\mathit{sec}}$ is defined on 188 documents submitted in yearly fashion to the U.S. Securities and Exchange Commission (SEC) by publicly traded companies, company insiders, and brokers\footnote{\it Company Filings: https://www.sec.gov/edgar}. Table~\ref{tbl:tasks} provides information about corpus associated with each task. Further details can be found in Appendix~\ref{sec:task-analysis}.

We selected these four tasks to cover a broad spectrum of knowledge domains, ranging from technical operations and community-driven Q\&A platforms to financial earnings and academic research, ensuring a diverse and comprehensive coverage of subjects. 

\subsection{RAG Pipelines}
\label{sec:ragpipelines}
 In our experiment benchmark, we consider $45$ different RAG pipelines by combining $5$ different retrieval mechanisms, $3$ different LLMs, and $3$ different ICL modes.

\noindent \textbf{Retrieval Mechanism Variants.} We consider the following $3$ retrieval paradigms: Closed-Book, Classical Retrieval, and Oracle. Closed-Book and Oracle act as lower and upper bounds on the quality of the information that can be provided to the LLM from the corpus. We also introduce five different classical retrieval methods, totalling seven retrieval mechanisms.

\noindent \textit{Closed-Book Retrieval.} No additional knowledge from the document corpus is provided to the LLM through retrieval. The exam-taker has only access to the question and the possible answers as well as the knowledge encoded in the weights of the LLM (i.e., parametric knowledge). We denote this method as \texttt{ClosedB}. A good evaluation score in this case relates to the LLM base knowledge of the question. Low \texttt{ClosedB} evaluation scores convey that the pre-trained model knows little about the domain or that the question or its possible answers are poorly formulated. 

\noindent \textit{Oracle.} The exam-taker has access to the specific document used to generate the question and answer pair, in addition to the question itself and all possible candidate answers. In other words, the exam-taker has access to the ground truth knowledge. A good \texttt{Oracle} score relates not only to the LLM base knowledge of the question, but also the ability to extract the answer from the ground truth. High \texttt{Oracle} scores can be obtained if the questions are properly formulated and the exam-taker is competent enough to extract the information to correctly answer. The \texttt{Oracle} score is uniquely possible thanks to our exam-design strategy and is core to providing a calibrated evaluation metric.

\noindent \textit{Retrieval Models.} The exam-taker is allowed to search over the knowledge corpus to combine the contextual knowledge with its parametric knowledge, using a given retrieval algorithm to better inform its answer. To give a representative perspective of the space of retrieval models, we compare a variety of methods.

\begin{itemize}
    \item \textit{Dense models:} We focus on two models: \texttt{MultiQA} embeddings \cite{talmor2019multiqa, minilm} and Siamese network embeddings (\texttt{SIAM}) \cite{koch2015siamese}.
    \item \textit{Sparse models:} We focus on \texttt{BM25} \cite{robertson2009probabilistic}, a widely-used information retrieval technique which employs a probabilistic model to rank documents based on the frequency and distribution of query terms within them.
    \item \textit{Hybrid models:} We consider ensembles of Dense and Sparse base retrievers where the output is re-ranked using a cross-encoder model~\cite{yadav2022efficient}. We refer to the models as \texttt{DPR} (\texttt{SIAM} plus \texttt{BM25}) and \texttt{DPRV2} (\texttt{MultiQA} plus \texttt{BM25}) bellow. 
\end{itemize}

Our analysis covers a spectrum of retrieval models, including contemporary models like \texttt{MultiQA} from Sentence Transformers and Cross-encoders in \texttt{DPR} and \texttt{DPRV2}, which are among the most used in the community (resp. 1.6M and 1.3M monthly downloads on HuggingFace, at the time of publication). BM25 is a standard bearer in Information Retrieval known for its robustness over modern methods. Our set of models is a combination of dense, sparse and hybrid models to ensure that our results are representative of all main classes. 

\noindent \textbf{LLM Variants.} We employ Mistral-7B, LlamaV2-13B and LlamaV2-70B ~\cite{jiang2023mistral, touvron2023llama}. We chose these three LLMs with the objective of investigating the spectrum of performance across different scales, aiming to gain insights into how the size of a model influences its language processing capabilities. These models offer a balance between advanced features, optimal performance at the time of the publication, community support, and practical considerations like resource availability and computational efficiency. Our original analysis also considered LlamaV2-7B, Falcon-40B and Alpaca-13B, which we discarded as they were consistently outperformed.

Finally, we consider the following $3$ in-context demonstration modes: ICL@0, ICL@1, and ICL@2. In the former, no in-context example is added to the prompt. In the two others, respectively one and two examples are provided in the prompt (question, candidate answers and correct one). While we examine performance on these specific RAG settings, our broader goal is not to maximize metrics on any single RAG formulation, but rather to have an evaluation system that is adaptable and extensible. 
Acknowledging the high frequency of top models release, our exam-based framework is intentionally crafted to be independent of the choice of retrieval approach or LLM, as discussed in Section~\ref{sec:exam-eval}. On a per-application basis, our approach allows to flexibly incorporate additional RAG dimensions like data processing methods, query reformulation, fine-tuning, etc.

\section{Experiment Results for Model Evaluation}
\label{sec:exp}
In this section, we present the experimental results for model evaluation by following the methodology discussed in Section~\ref{sec:model} and the benchmark introduced in Section~\ref{sec:benchmark}. We first introduce point-wise evaluation results of RAG pipelines in Table~\ref{tab:accuracy} and then discuss IRT-based ability levels for individual RAG components in Table~\ref{tab:ability}. Such results are used at task level to make optimal design decisions and across tasks to infer RAG system patterns.

\begin{table}[t]
\centering
\small
\begin{tabular}{clccccc}
\toprule
\multicolumn{6}{c}{\textbf{Best Absolute Accuracy in \%}} \\ \hline
 & \textbf{Retrieval} & $t_{\mathit{ops}}$ & $t_{\mathit{stk}}$ & $t_{\mathit{arx}}$ & $t_{\mathit{sec}}$ & \textbf{Avg.} \\ \hline
\multirow{7}{*}{\rotatebox[origin=c]{90}{Mistral-7B}}
&\texttt{ClosedB}& 52.2	& 48.6 &	54.5 & 49.5 & 51.2 \\
&\texttt{SIAM}& 45.5	& 50.0 &	57.0 & 47.6 & 50.0  \\
&\texttt{DPR}& 52.2	& 58.3 &	60.3 &60.5 & 57.8 \\
&\texttt{BM25}& \textbf{58.0}	& 60.4 &	69.5 & 55.3 & 60.8 \\
&\texttt{MultiQA}& 57.7	& \textbf{72.2} &	\textbf{69.5} & 53.6 & 63.2 \\
&\texttt{DPRV2}& 55.1	& 70.1 &	69.4 & \textbf{63.9} & 64.6 \\
&\texttt{Oracle}& 63.8	& 74.3 &	68.6 & 70.9 & 69.4 \\
\hline
\multirow{7}{*}{\rotatebox[origin=c]{90}{LlamaV2-13B}}
&\texttt{ClosedB} & 50.4	& 42.4	& 45.5 & 45.8 & 46.3 \\
&\texttt{SIAM} & 44.6	& 48.6	& 56.2 & 42.7 & 48.0 \\
&\texttt{DPR} & 51.3	& 54.9	& 55.4 & 61.0 & 55.7 \\
&\texttt{BM25} & 54.5	& 63.2	& 66.9 & 55.9 & 60.1 \\
&\texttt{MultiQA} & \textbf{58.0}	& 67.4	& 66.1 & 53.8 & 61.3 \\
&\texttt{DPRV2} & 55.7	& \textbf{66.7}	& \textbf{69.4} & \textbf{63.9} & 63.9 \\
&\texttt{Oracle} & 63.0	& 68.8	& 68.6 & 69.9 & 67.6 \\
\hline
\multirow{7}{*}{\rotatebox[origin=c]{90}{LlamaV2-70B}}
&\texttt{ClosedB}&	63.0 & 38.2 & 	47.1 & 48.7 & 49.3 \\
&\texttt{SIAM}& 	55.4 & 53.5 & 	57.9 & 49.5 & 54.1 \\
&\texttt{DPR}&	61.5 & 59.7 & 	60.3 & \textbf{68.3} & 62.5 \\
&\texttt{BM25}&	71.4 & 65.3 & 	76.9 & 64.3 & 69.5 \\
&\texttt{MultiQA}&	\textbf{72.6} & \textbf{75.7} & 	74.4 & 58.6 & 70.3 \\
&\texttt{DPRV2}&	69.7 & 72.2 & 	\textbf{77.7} & 67.6 & 71.8 \\
&\texttt{Oracle}&	72.6 & 73.6 & 	76.0 & 77.1 & 74.8 \\
\bottomrule
\end{tabular}
\caption{Point-wise evaluation results. The score is the percentage of correctly answered questions by the RAG. More precisely, we denote the maximum score among the three ICL passes as 
\textit{best absolute accuracy} for a RAG. For each LLM, we indicate the top performing retriever in bold.}
\label{tab:accuracy}
\end{table}

\subsection{Accuracy and Ability Analysis }
\label{sec:pointwise}

In summary, our experiments result in the following four findings: Firstly, there’s no one size that fits all, i.e., the optimal choice of retrieval method, and to a lesser extent LLM, is typically task-dependent. Depending on the task and retrieval, Mistral-7B and LlamaV2-13B ranking varies. LlamaV2-70B is even outperformed in no-retrieval settings. Similarly, for some tasks such as $t_{\mathit{sec}}$ and $t_{\mathit{arx}}$, \texttt{BM25} outperforms \texttt{MultiQA} and \texttt{SIAM}, which indicates that sparse retrieval is typically better than dense retrieval for these tasks. A conjecture is that such tasks often contain easily identifiable terms (e.g., AWS service names in $t_{\mathit{ops}}$) which can be retrieved with keyword search, while other tasks like $t_{\mathit{stk}}$ mostly contains common words. However, our findings do indicate that hybrid ensemble models, which integrate both dense and sparse retrieval techniques (e.g., \texttt{DPRV2}), generally offer greater robustness and adaptability across a variety of tasks compared to exclusively dense or sparse models. 

Secondly, the right choice of the retrieval method can often lead to performance improvements surpassing those from simply choosing larger LLMs, as seen when comparing marginal gains in Table~\ref{tab:ability}: in $t_{\mathit{sec}}$, we gain more ability gain by switching from \texttt{SIAM} to~\texttt{DPRV2} compared to switching to larger LLMs. Thirdly, for tasks involving \textit{closed source knowledge}, the accuracy bottleneck is typically the LLM and not the retrieval method.By \textit{closed source}, we refer to confidential data, proprietary to companies, such as internal financial statements, proprietary codebase, internal FAQs or documents. This type of corpus is particularly relevant given that the LLM wasn’t exposed to it during the pre-training: all the information flows through the retrieval. Fourthly, poorly aligned retriever component can lead to a worse accuracy than having no retrieval at all, as seen for \texttt{SIAM} performance compared to \texttt{ClosedB} in Tables~\ref{tab:accuracy} and~\ref{tab:ability}.

Finally, a noteworthy phenomena in RAG systems is when there is strong information overlap between documents. Notably, this explains why the Oracle might be outperformed by some retriever, as seen in Table~\ref{tab:accuracy} for $t_{\mathit{arx}}$ and in Figure~\ref{fig:all_radar}: certain document chunks are more helpful to answer question than the one used to actually generate the question.

\begin{table}[t]
\centering
\small
\begin{tabular}{cccccc}
\toprule
\multicolumn{6}{c}{\textbf{Ability Level}} \\ \hline
 & \textbf{Retrieval} & $t_{\mathit{ops}}$ & $t_{\mathit{stk}}$ & $t_{\mathit{arx}}$ & $t_{\mathit{sec}}$  \\ \hline
\multirow{3}{*}{\rotatebox[origin=c]{90}{LLM}}
&Mistral-7B  & -0.38 & -0.59 & -1.03 & -0.48 \\ 
&LlamaV2-13B & -0.04 & -0.51 & -0.78 & -0.36 \\ 
&LlamaV2-70B & 1.00 &  0.40 & -0.05 & 0.18 \\ 
\hline
\multirow{7}{*}{\rotatebox[origin=c]{90}{Retrieval}}
&\texttt{ClosedB}  & -0.86 & -1.39 & -0.62 & -1.29 \\
&\texttt{SIAM}        & -2.74 & -0.06 & -0.21 & -1.39 \\
&\texttt{DPR}         & -0.87 &  0.36 & -0.01 & 0.10 \\
&\texttt{BM25}        & -0.56 &  0.62 &  0.60 & -0.22 \\
&\texttt{MultiQA}     & -0.43 &  1.06 &  0.62 & -0.42 \\
&\texttt{DPRV2}       & -0.54 &  0.99 &  0.72 & 0.22 \\
&\texttt{Oracle} & -0.20 & 1.14 &  0.75 & 0.59 \\ 
\hline
\multirow{3}{*}{\rotatebox[origin=c]{90}{ICL}}
&ICL@0       & -0.54 & -0.77 & -0.11 & -0.83 \\
&ICL@1       &  0.66 & 0.02 &  0.90 & 0.05 \\
&ICL@2       &  0.46 & 0.04 &  1.06 & 0.11 \\
\bottomrule
\end{tabular}
\caption{IRT evaluation results for each RAG component $(\theta_{\mathit{llm}(m)}, \theta_{\mathit{ret}(m)}, \theta_{\mathit{icl}(m)})_{m\in\mathcal{M}}$. A higher level of model ability level corresponds to a higher value of $\theta$, and values are relative: for instance, to assess the ability gain of a given retrieval model MultiQA, we consider $\theta_{\texttt{MultiQA}}-\theta_{\texttt{ClosedB}}$. Note that results are not normalized across tasks and thus not directly comparable. See Table~\ref{tbl:irt_fit} for question-based parameters $(g_{i}, b_{i}, d_{i},)_{i\in \mathcal{M}}$.}
\label{tab:ability}
\end{table}

\subsection{Evaluating the Evaluation: Meta-Evaluation}
\label{sec:metaevaluation}

In Section~\ref{sec:pointwise}, we presented how our evaluation framework assesses various RAG pipelines by utilizing point-wise evaluation results and IRT-based ability levels for individual RAG components. Another critical question is how to evaluate our evaluation framework itself.

Comparing and evaluating the evaluation methods of LLMs, including our exam-based evaluation model, is a complex meta-evaluation task. Granted current challenges for direct evaluation of LLMs, we highlight that performing meta-evaluation is a step above in terms of difficulty. Beyond that, meta-evaluation of LLM assessment is a multi-objective problem due to the multidimensional nature of LLM performance: LLMs are assessed on varied capabilities like factuality, linguistic understanding, coherence, and ethical considerations, each requiring specific evaluation criteria. The rapid evolution of LLM technology adds to this complexity, as new models may exhibit behaviors not previously considered, necessitating continuous updates to meta-evaluation methodologies.
Furthermore, the subjective nature of language processing and the diversity of LLM applications demand different performance metrics, further complicating the meta-evaluation process. The reliance on human judgment as a benchmark introduces variability, making it challenging to establish a universal evaluation framework that balances technical accuracy with diverse human perspectives and real-world applicability~\cite{howcroft2020twenty}. 

Typical NLP evaluation methods like ROUGE, BLEU, and BERTScore, commonly used for evaluating specific aspects of language models, are too narrow for effectively meta-evaluating LLMs, lacking breadth, interpretability, and feedback to assess capabilities and guide improvements. Recent LLM-based evaluation methods~\cite{es2023ragas,fu2023gptscore,zheng2023judging,xu2023interactive} are promising but still have limitations in scope, adaptability, interpretability, bias reduction, or actionable feedback required for comprehensive LLM evaluation. A key distinction of our exam-based evaluation approach compared to other methods is that it is interpretable and provides predictive and prescriptive guidance on areas where the RAG needs improvement.

\section{Experiment Results for Exam Evaluation}
\label{sec:examresults}

Properly defining what is a good exam is a difficult question: although perfectly correct from a content perspective, an exam can still be of lower quality by not being discriminative enough across models nor informative enough on the task of interest.
To quantitatively measure and improve upon this, we present in this section an analysis of the exam questions generated by our framework across different categorization schemes. Specifically, we leverage Bloom's taxonomy to categorize questions by cognitive complexity and introduce an \textit{item information function} to quantify the informativeness of questions for evaluating model performance. Figures~\ref{fig:bloom} and~\ref{fig:seman} illustrate this process in the context of StackExchange task. We conclude by presenting a methodology to iteratively maximize the informativeness of the exam, a key contribution of our work. 

\subsection{Exam Informativeness}
\label{sec:exam-informativeness}

To measure the informativeness of the exam with respect to the task and models, we introduce the \textit{item information function}, aka, Fischer information~\cite{hambleton1991fundamentals}. This function quantifies the amount of information that an observable random variable $X$ provides about the unknown ability parameter $\theta$, through a measure of the curvature of the log-likelihood function $\mathcal{L}$. Thereby, it offers a pivotal metric for assessing the precision of statistical estimators in parameter estimation theory and more precisely the discriminating power of the exam question over the space of candidate models at different levels of ability. It is defined for an individual question as:
\begin{equation}
    I(\theta|g_i, d_i, b_i) = d_{i}^{2}\frac{(p_{i}(\theta)-g_{i})^{2}}{(1-g_{i})^{2}}\frac{1-p_{i}(\theta)}{p_{i}(\theta)},
\end{equation}
In Figure~\ref{fig:irt_sec}, we highlight individual item information functions for this task. Note that the item information function reaches its maximum value at the question's difficulty parameter. Thus, questions provide the most information for estimating $\theta$ at an ability level close to their difficulty, and provide less information at ability levels further away from their difficulty. In this way, the item information function formally characterizes a question's capacity for discriminating between individuals and around a particular ability level. To asses the overall effect of a given subset of questions $\mathcal{R} \subset \mathcal{Q}$, we introduce the aggregated Information function:
\begin{equation}
\Bar{I}_{\mathcal{R}}(\theta) = \frac{1}{|\mathcal{R}|}\sum_{i\in \mathcal{R}} I(\theta|g_i, d_i, b_i),
\end{equation}

\begin{figure}[t]
    \centering
    \includegraphics[width=\linewidth]{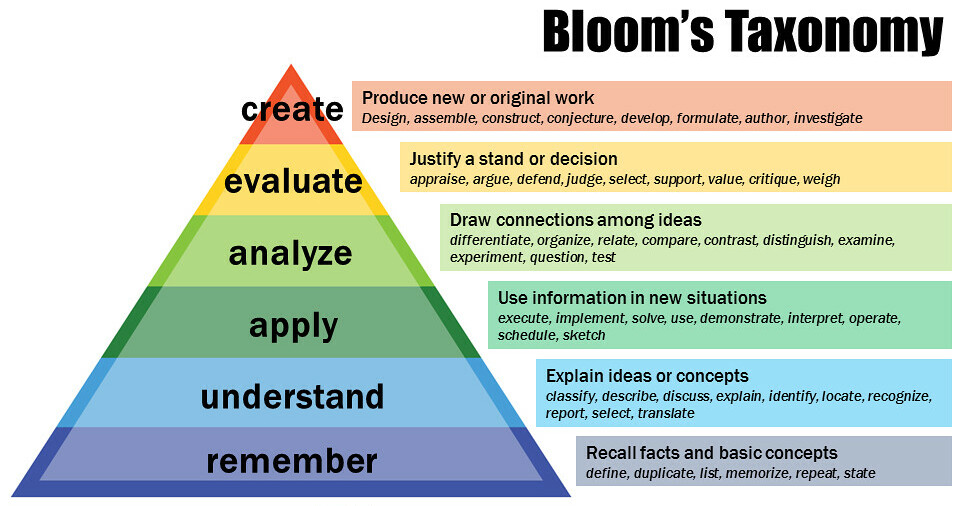}
    \caption{Representation of Bloom's revised taxonomy. The cognitive complexity of skills increase from the bottom to the top of the pyramid. Source: \citealp{bloom}}
    \label{fig:bloom-pyramid}
\end{figure}

\subsection{Categorization of Exams and Questions}
\label{sec:cat-exam-qna}

Once an exam is generated, we perform an automated question categorization to determine the relevant dimensions for a given question. Question categories enable a more granular understanding of types of questions that RAG pipelines are better or worse at as well as the ones that helps to better discriminate across models, through the usage of the item information function introduced above. For this aim, we leverage Bloom's revised taxonomy \citep{bloom1956taxonomy, krathwohl2002revision} illustrated in Figure \ref{fig:bloom-pyramid}. Bloom's taxonomy is a hierarchical model that classifies learning objectives into different levels of cognitive complexity. Table~\ref{tbl:bloom} in Appendix~ \ref{sec:bloom} illustrates the levels of the revised Bloom's Taxonomy, from the lowest to the highest, along with a brief description and examples of how they might translate into multiple-choice questions. They differentiate between the knowledge dimension (factual, conceptual, procedural, and meta-cognitive) and the cognitive process dimension (remember, understand, apply, analyze, evaluate, and create).

\begin{figure}[t]
    \centering
    \includegraphics[width=\linewidth]{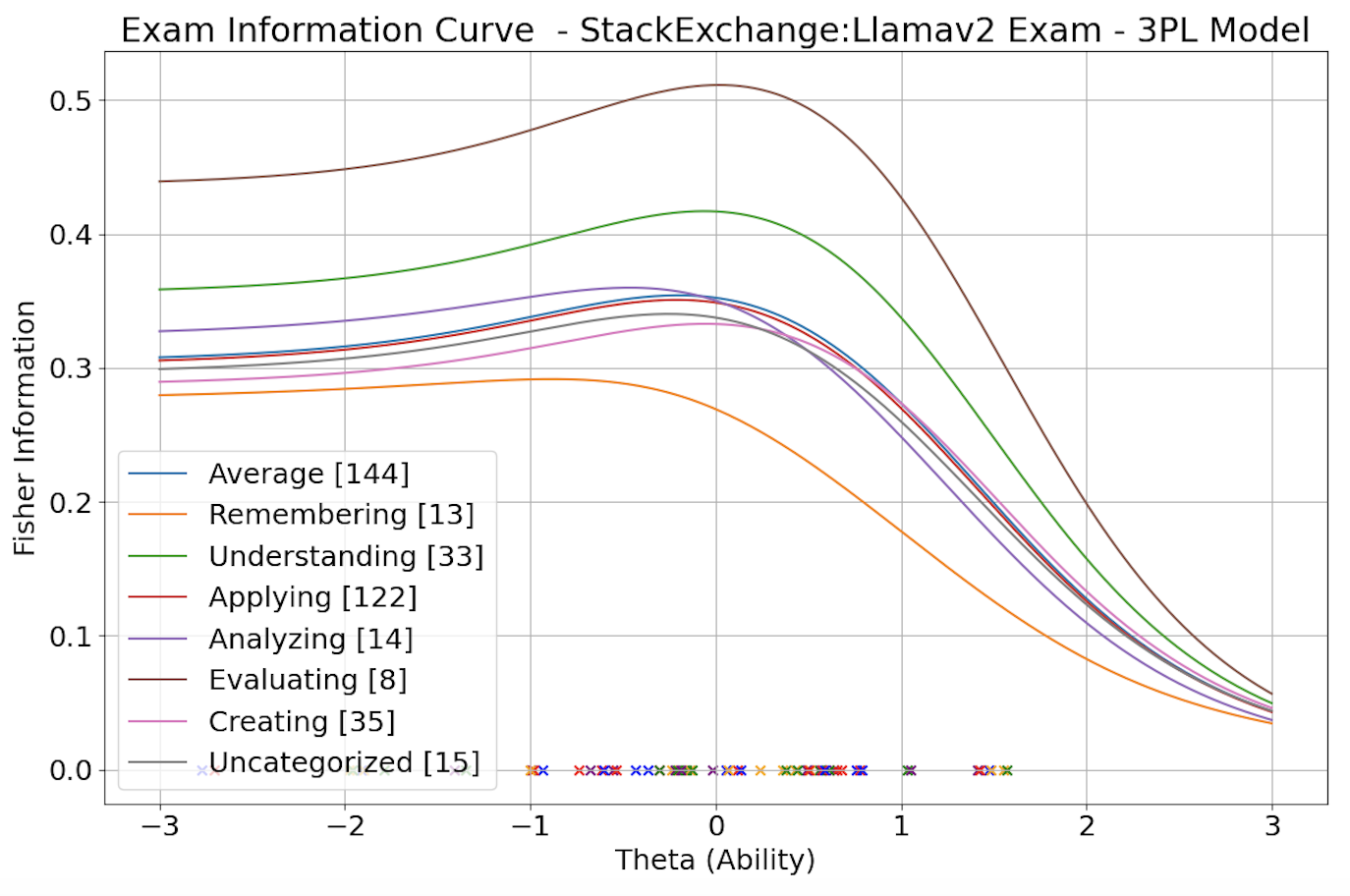}
    \caption{Aggregated Information function $\Bar{I}_{cat}(\theta)$ for $t_{\mathit{stk}}$, averaged according to Bloom taxonomy. Each cross on the x-axis correspond to a given model ability $\theta_{m}$, with no particular signification granted to their colors.}
    \label{fig:bloom}
\end{figure}

In Figure \ref{fig:bloom}, we present the average item information function $\Bar{I}_{cat}(\theta)$ for each Bloom category for $t_{\mathit{stk}}$. Lines are fitted by maximizing the log-likelihood $\mathcal{L}$ defined in Equation ~\ref{eq:log-irt}, using the optimization process described in Appendix ~\ref{sec:hierarchical-irt-fit}. Informativeness is an increasing quantity, meaning that higher values are better. As discussed in Section~\ref{sec:exam-eval}, it is also a function of the ability level. Therefore, some questions might be more informative at certain level of ability, for instance to discriminate among medium ability students and less at others, such as for high ability students. For this specific task $t_{\mathit{stk}}$, we observe that \textit{evaluating} and \textit{understanding} are the most discriminate dimensions in Bloom's taxonomy across different ability levels, where \textit{remembering} is the least discriminatory. Such task-specific insights  empower the decision-makers to better evaluate and understand the task, and highlight model strengths and limitations. 

\begin{figure}[t]
    \centering
    \includegraphics[width=\linewidth]{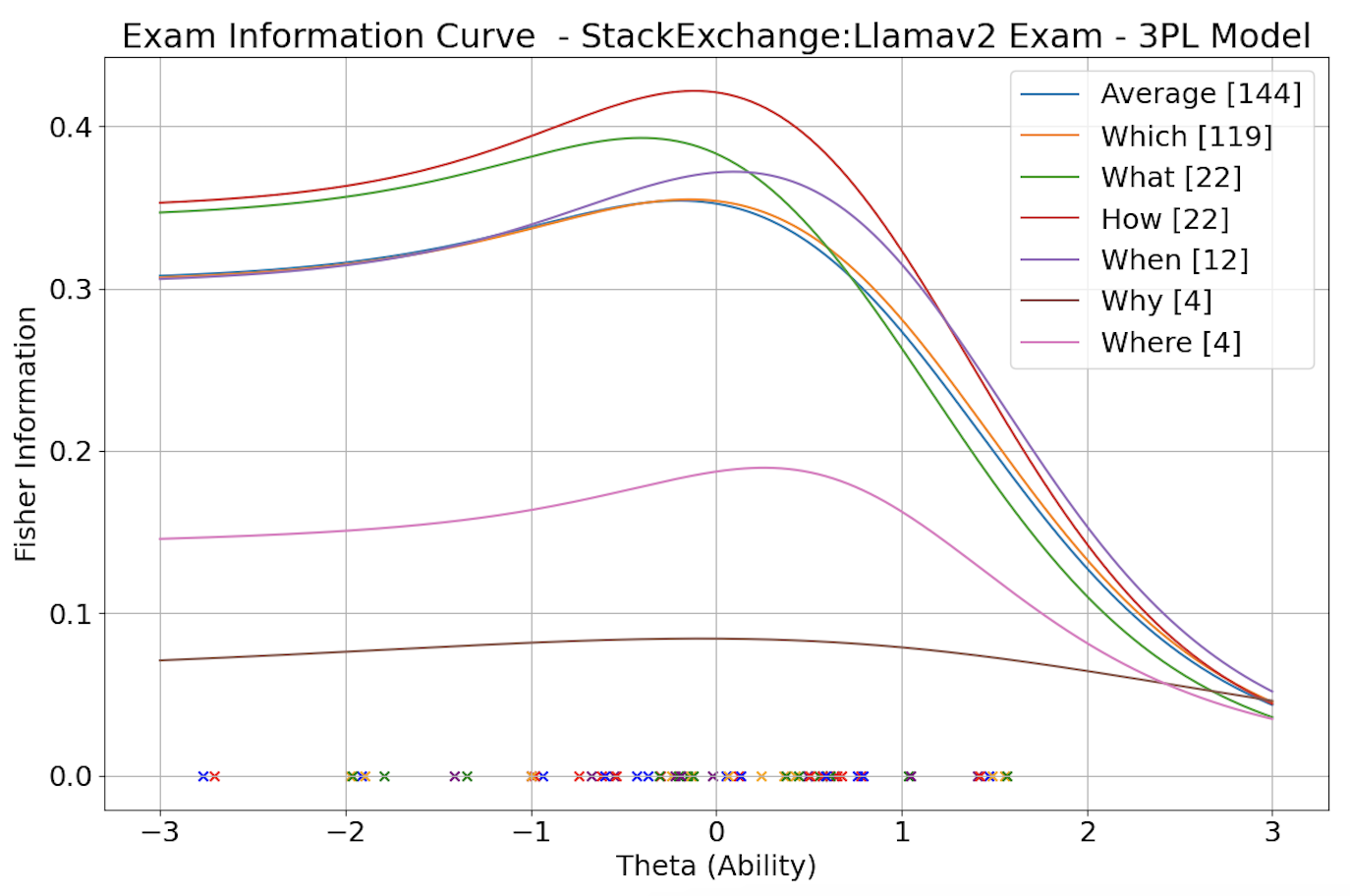}
    \caption{Aggregated Information function $\Bar{I}_{cat}(\theta)$ for $t_{\mathit{stk}}$, averaged according to semantic taxonomy. Each cross on the x-axis correspond to a given model ability $\theta_{m}$.}
    \label{fig:seman}
\end{figure}

Similarly, Figure~\ref{fig:seman} shows a clustering of questions in the task $t_{\mathit{stk}}$ based on semantic type (e.g., where, what). We observe that \textit{What} and \textit{Which} were the most discriminatory for lower ability levels, and \textit{When} discriminated more at higher ability levels. One interpretation is that \textit{What} and \textit{How} questions tend to be more factual and syntax-based in the $t_{\mathit{stk}}$ domain, and hence RAG with lower ability level struggle more with these genres of questions. \textit{When} question may also involve more situational logic where RAG with higher ability level are better equipped to answer. We refer the reader to Appendix~\ref{sec:hierarchical-irt} for further discussions and extended analysis of the Hierarchical IRT Model on the other tasks.

Moreover, we argue that our novel programmatic application of Bloom’s Taxonomy and Item Response Theory provides a more comprehensive understanding of the exam's framework, thereby aiding practitioners in identifying potential biases. Specifically, Figure~\ref{fig:bloom} showcases the distribution of question types (e.g., where, what, who), Figure~\ref{fig:seman} details the taxonomy of questions (e.g., creating, evaluating, remembering...), and Table~\ref{tbl:exam_stats} offers key statistics on questions and answers. Together, these elements offer fresh insights into the exam structure and contribute significantly to the identification and mitigation of biases.

\subsection{Iterative Exam Improvement}
\label{sec:iterative-improvement}

Lastly, in order to increase the quality of the exam and thus better distinguish among the highest performing RAG pipelines, we introduce an iterative method to generate new exams $\mathcal{Q}_{1} \mapsto \mathcal{Q}_{2} \dots \mapsto  \mathcal{Q}_{n}$ by adaptively selecting questions to maximize the informativeness: $\Bar{I}_{\mathcal{Q}_{1}}(\theta) \preceq \Bar{I}_{\mathcal{Q}_{2}}(\theta) \dots \preceq \Bar{I}_{\mathcal{Q}_{n}}(\theta)$. More precisely, we apply an alternate process of IRT model fitting and question discarding based on the inferred discrimination parameter $(d_{i})_{i\in \mathcal{Q}}$. The methodology is discussed in details in Appendix~\ref{sec:iterative-irt}. Figure.~\ref{fig:iterative-improvemenet-exam} illustrate the maximization process for $t_{\mathit{arx}}$ as the exam and IRT estimation evolves; other tasks are presented in Appendix~\ref{sec:iterative-irt}. For $t_{arx}$ or $t_{ops}$, we witness a continuous Pareto-dominating improvement, although mostly in the low to medium ability levels: the exam becomes more and more informative with the iterations. Such improvement is also witnessed for $t_{stk}$, although convergence happens faster. Finally, for $t_{sec}$, the evolution is non-monotonic and interestingly mostly happens in high ability regions. To conclude, this process is the first step towards a data-driven continuous optimization of the exam and we believe it is one of the most promising follow-up direction for the field of automated evaluation.

\begin{figure}[t]
    \centering
    \includegraphics[width=\linewidth]{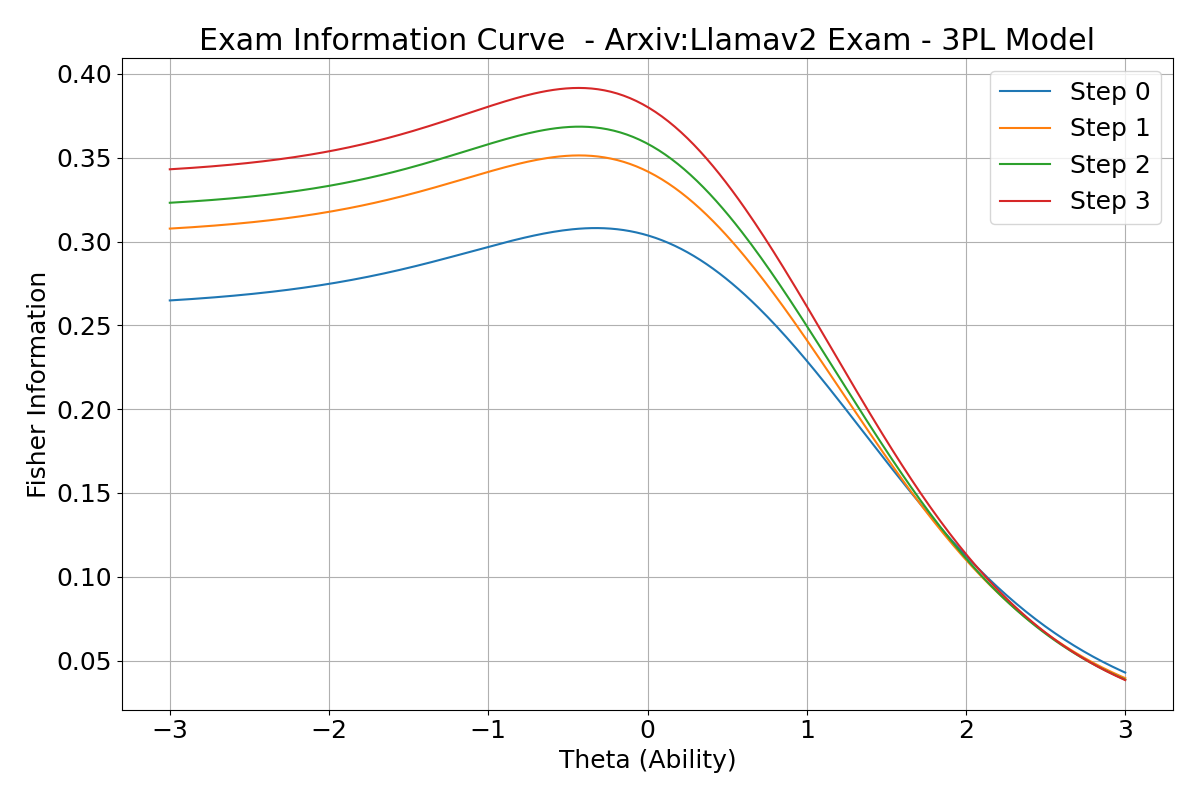}
    \caption{Evolution of Exam Informativeness during the iterative process, for ArXiv task. Each curve represents the exam aggregated Information function $\Bar{I}_{\mathcal{Q}_{i}}(\theta)$, at step $i$.}
    \label{fig:iterative-improvemenet-exam}
\end{figure}
\section{Conclusion}
\label{sec:conc}

In this paper, we proposed and demonstrated a robust method to evaluate the performance of Retrieval-Augmented Large Language Models on specific tasks. By automatically generating multiple choice exams tailored to the document corpus associated with each task, our approach enables standardized, scalable, and interpretable scoring of different RAG systems. Through iterative optimization guided by Item Response Theory, we create highly informative exams that surface the strengths and weaknesses of different model configurations. Our experiments on question answering across four distinct domains reveal key insights into the factors driving RAG performance. Notably, we find that optimization of the retrieval mechanism can unlock bigger gains than simply scaling model size, highlighting the importance of a co-design approach. Overall, our work provides an efficient, reproducible paradigm for benchmarking and improving RAG for real-world applications. 

Natural extensions of our work include investigating multi-language applications, incorporating agent-based systems for sequential decision-making tasks to extend beyond RAG systems, and utilizing the exam-based approach in traditional NLP problems like summarization and translation, thereby fostering the creation of more nuanced benchmarking datasets.




\newpage

\section*{Impact Statements}
\label{sec:societal}


This paper presents work whose goal is to advance the field of Machine Learning. There are many potential societal consequences of our work, none which we feel must be specifically highlighted here. While enhanced factual accuracy for language models could have broad positive applications, we acknowledge there may also be risks if used improperly.

\bibliography{rag-eval}
\bibliographystyle{icml2024}

\newpage
\appendix
\onecolumn

\section{Details on Exam Generation}
\label{sec:exam-details}

\begin{figure}[!h]
\includegraphics[width=\textwidth]{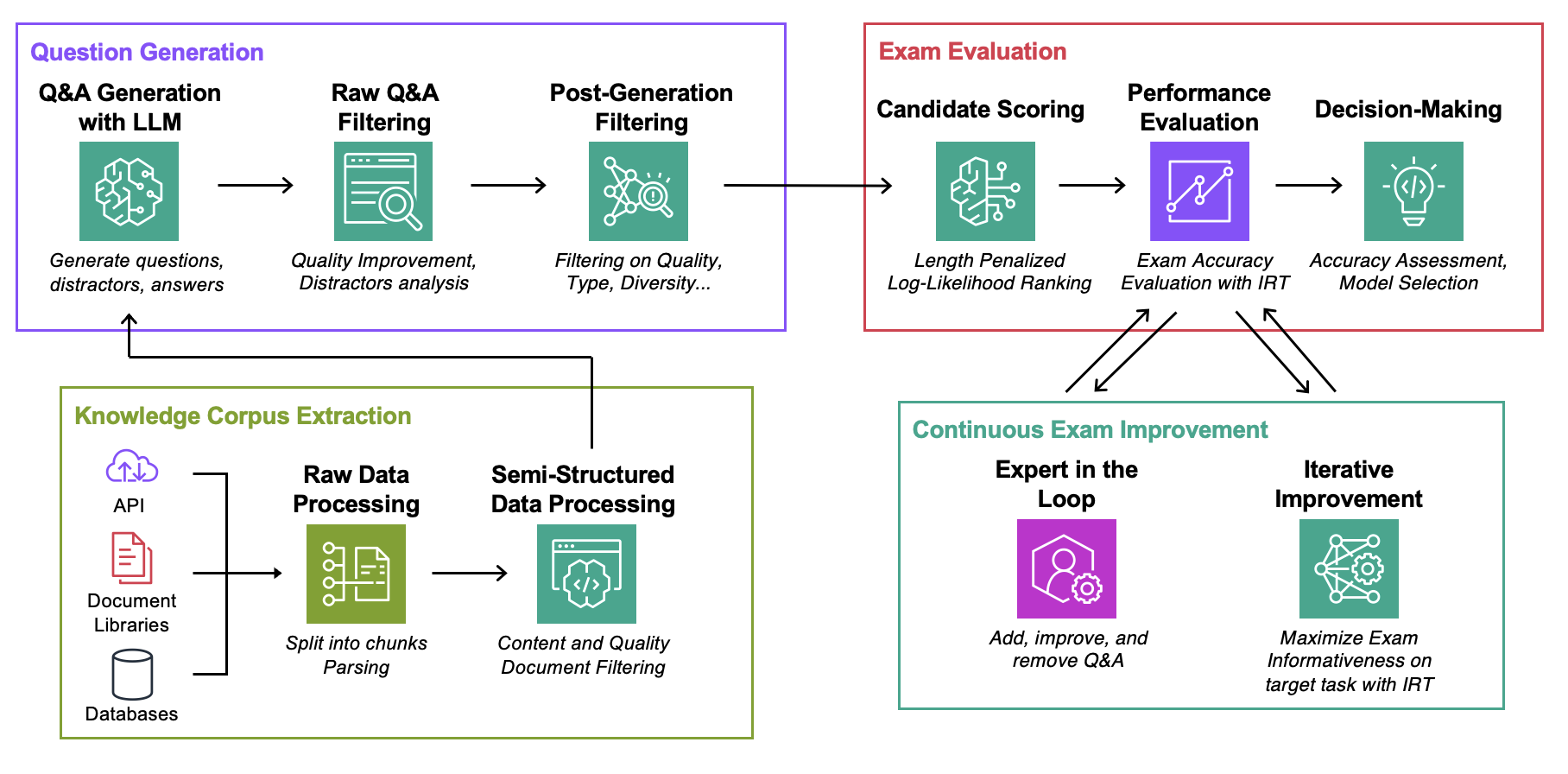}
\caption{Summary of the exam generation, evaluation and iterative improvement processes.} 
\label{fig:summary_sketch}
\end{figure}

In this section, we present in details the exam generation process. Figure~\ref{fig:summary_sketch} is provided as a summary and includes an overview of the exam generation, evaluation and continuous improvement.The exam generator algorithm uses a pre-trained LLM to generate a multiple choice exam with $n$ questions on a specific task $t \in \mathcal{T}$. In this work, we relied on LLamaV2-70B to generate the questions, after a preliminary comparative study with LlamaV2-13B, ClaudeInstant and ClaudeV2. We differ an extensive analysis of the difference to follow-up work. As shown in Figure~\ref{fig:summary_sketch}, the algorithm has a three-step approach: first it generates candidate questions and answers pairs from a subset of all documents, either random or topic selected. It then filters the raw questions to remove low quality ones, notably by improving the quality of incorrect ``discriminator'' answer choices. Finally, it filters the correct questions to ensure maximal quality and add potential constraints on diversity. In this section, we illustrate the process used for exam generation (Section~\ref{sec:prompt-exam-generation}) and question filtering (Section~\ref{sec:question-filtering}). We present statistics on each exam in Section~\ref{sec:exam-stats-sec}. Next, in Section~\ref{sec:task-analysis}, for each of the four tasks considered in the paper, we describe the task and exam specificity. Finally, we conclude by presenting the granular accuracy results for each task in Section~\ref{sec:granular-results}.

\subsection{Prompt for Exam Generation}
\label{sec:prompt-exam-generation}
There are two variables in the following prompt: \texttt{task\_domain} and \texttt{documentation}. The former is one of the task names: DevOps, ArXiv, StackExchange, SEC Filings. The latter is the support documentation for the task.

\begin{promptexample}
Human: Here is some documentation from {task_domain}: {documentation}. From this, generate 
a difficult multi-form question for an exam. It should have 4 candidates, 1 correct answer
and explanations. 

Syntax should be:
Question: {question}
A) {candidate A}
B) {candidate B}
C) {candidate C}
D) {candidate D}
Correct Answer: {correct answer}

Assistant:
\end{promptexample}

\subsection{Question Filtering}
\label{sec:question-filtering}

\paragraph{Raw Question Parsing:} To correctly extract the answer from the model generated answer, we use regular expression filters for both the question, the candidate choices and the correct answer. If one of the parsing fails, we discard the question. See Table~\ref{tbl:exam_stats} for some statistics on parsing efficiency.

\paragraph{Candidate Shuffling:} We randomly shuffle the candidate answers order to ensure that random guessing performance is around 25\%. This prevents a model bias of choosing the first candidate as the correct answer, as observed around $40\%$ of the time during our experiments. See Table~\ref{tbl:exam_stats} for the baseline performance of $(i)$ picking the same fixed answer for all questions, and $(ii)$ always picking the longest answer. Aside from these two, we didn't detect any notable bias in terms of correct answer generation.

\paragraph{Self-contained Constraint:} We use some regular expression checks to ensure that the question is self-contained, i.e. that is doesn't contain explicit references to the documentation used to generate it. In particular, we filter out questions with the following patters:
\begin{promptexample}
# term immediately followed by title in quotes
r'\b(documentation|paper|article|research|study)\b\s*\"[^\"]+\"'

# citation-like sentence followed by title
r'\b(discussed in|addressed in|described in|of the)\b\s*\"[^\"]+\"'

# fallback to original terms
r'\b(documentation|paper|article|research|study)\b'
\end{promptexample}

\paragraph{Discriminator Analysis:} As mentioned in Section~\ref{sec:exam-gen}, we note an interesting asymmetry: granted a document corpus, it is relatively easy for an LLM to generate a question and the correct answer, as this task is self-contained in the prompt in term of knowledge. However, it is considerably more difficult to create high quality incorrect answers, commonly refereed as discriminators.  One bias of the model is to create either multiple rephrased correct answers or to create discriminators, which are correct but incomplete. To filter such questions out, for each question, we use Jaccard similarity at $n$-gram level $\mathcal{J}_{n}$ (where $n$ is picked as the mean candidate answer token length), as well as embedding similarity $\mathcal{S}$. Given an exam question $q\in \mathcal{Q}$ based on documentation $k$, the correct answer $c$ and the discriminators $d_{i}$, we introduce two filters:
\begin{itemize}
    
    \item \textbf{Extra-Candidate Similarity}: If $\mathcal{J}(k, c) + t_{1} < \max_{i} \mathcal{J}(k, d_{i})$, or $\mathcal{S}(k, c) + t_{2} < \max_{i} \mathcal{S}(k, d_{i})$, for $t_{1}, t_{2}$ two threshold values, we remove the question $q$ from the corpus. This check ensures that discriminators are not closer in meaning to the documentation compared to the original question. In practice, we obtain the best results by setting values of $t_{1}, t_{2}$ such that around $5\%$ of questions are removed.
    \item \textbf{Intra-Candidate Similarity}: If $\max_{i} \mathcal{J}_{n}(c, d_{i}) \geq t_{3}$ or $\max_{i} \mathcal{S}(c, d_{i}) \geq t_{4}$, for $t_{3}, t_{4}$ two threshold values, we remove the question $q$ from the corpus. This check ensures that discriminators are not too close to the original question. In practice, we obtain the best results by setting values of $t_{3}, t_{4}$ such that around $5\%$ of questions are removed.
\end{itemize}



\subsection{Generated Exam Statistics}
\label{sec:exam-stats-sec}

\begin{table*}[!h]
\centering
\begin{tabular}{lrrrr}
\toprule
\textbf{Attribute} & $t_{\mathit{ops}}$ (\textbf{DevOps}) & $t_{\mathit{arx}}$ (\textbf{Arxiv}) & $t_{\mathit{stk}}$ (\textbf{StackExchange}) & $t_{\mathit{sec}}$ (\textbf{SEC Filings}) \\
\midrule
\# Candidate Q\&A Generated       & $700$ & $500$ & $326$ &  $520$ \\
\# Incorrectly Parsed Q\&A       & $126$   & $119$   & $143$ & $35$ \\
\# Incorrect Discriminators/Self-Contained Q\&A   & $99$   & $32$   & $5$ & $41$\\
\# Surviving Q\&A   & $275$   & $381$ & $148$ & $515$\\
Fixed Answer Baseline Accuracy           & $27.6\%$   & $27.6\%$ & $29.7\%$ & $27.4\%$ \\
Longest Answer Baseline Accuracy    & $36.0\%$   & $37.3\%$ & $38.51\%$ & $41.6\%$\\
Avg. Question Length (character)     & $303$   & $355$ & $270$ & $141$\\
\bottomrule
\end{tabular}
\caption{Analysis of the exam generation process. We present both data on the exam generation and refinement process, as well as on the resulting exam characteristic.}
\label{tbl:exam_stats}
\end{table*}

 To edge against LLM biases in question generation, we compute in Table~\ref{tbl:exam_stats} two baselines. First, the fixed answer baseline assesses the score of always picking the same given answer across all questions. By randomising the candidate answers order during the exam generation, we ensure that this baseline accuracy is around $25\%$ as expected. Secondly, the longest answer baseline is the score obtained by a model always picking the longer answer. Although we observe a higher score than a random pick, the value is still reasonably low not to motivate a programmatic correction. No other potential biases of this type were detected during our analysis.

\subsection{Knowledge Corpus and Task Creation}
\label{sec:task-analysis}

\vspace{5pt}
\noindent \textbf{Task $t_{\mathit{ops}}$ (DevOps).} The task~$t_{\mathit{ops}}$ is defined over a knowledge corpus of 1249 webpages from AWS Knowledge Center \footnote{\it AWS Knowledge Center: https://repost.aws/knowledge-center.}  where each webpage troubleshoots one DevOps problem for AWS customers. Tags associated with the HTML code allows to cluster the questions on 18 overlapping topics. To create document chunks, we break down the documents in non-overlapping chunks to ensure a maximum of $10$ sentences per chunk, a character length between $500$ and $4500$ and a token length between $200$ and $900$. This results in a total of $4564$ document chunks. The following example is a question generated for an AWS cloud storage service called S3.

\begin{promptexample}
{
  "question": "You are an AWS engineer responsible for monitoring the storage usage of your company's Amazon S3 buckets. You want to track the total storage usage and number of objects in each bucket. Which of the following metrics in CloudWatch should you use to achieve this?",
  "documentation": "However, as soon as the objects are marked for deletion, you are no longer billed for storage (even if the object isn't removed yet). Note that the Amazon S3 monitoring metrics are recorded once per day. Therefore, these metrics might not display the most updated information. However, CloudWatch monitors your AWS resources and applications in real time . Also, S3 console and Storage Lens use base 2 conversion (/1024) to report storage metrics, and CloudWatch by default uses base 10 conversion (/1000). Resolution Daily storage metrics in CloudWatch In CloudWatch, the BucketSizeBytes metric captures all Amazon S3 and Amazon S3 Glacier storage types, object versions, and any incomplete multipart uploads. This value is calculated by summing up all object sizes, metadata in your bucket (both current and noncurrent objects), and any incomplete multipart upload sizes. For example, the BucketSizeBytes metric calculates the amount of data (in bytes) that's stored in an Amazon S3 bucket in all the following object storage classes : S3 Standard S3 Intelligent-Tiering S3 Standard-IA S3 One Zone-IA S3 Reduced Redundancy Storage S3 Glacier Deep Archive S3 Glacier Flexible Retrieval S3 Glacier Instant Retrieval Additionally, the NumberOfObjects metric in CloudWatch contains the total number of objects that are stored in a bucket for all storage classes. This value counts all objects in the bucket (both current and noncurrent), along with the total number of parts for any incomplete multipart uploads. The NumberOfObjects metric also calculates the total number of objects for all versions of objects in your bucket.",
  "choices": [
    "A) BucketSizeBytes",
    "B) ObjectVersionBytes", 
    "C) IncompleteMultipartUploads",
    "D) NumberOfObjects"
  ],
  "correct_answer": "A) BucketSizeBytes"
}
\end{promptexample}
Next, we present a random sample of 10 questions from the DevOps exam:
\begin{promptexample}
Question 1: You are an AWS engineer responsible for setting up a site-to-site VPN connection between your company's network and Amazon VPC. You have configured the VPN tunnel, but it is not establishing successfully. You have checked the AWS VPN configuration and found that it meets all the requirements mentioned in the documentation. However, you are still experiencing issues. Which of the following could be the cause of the problem?
Question 2: You are an AWS engineer responsible for managing an ECS cluster. You are receiving errors when trying to add tags to the cluster. The errors indicate that the IAM entity does not have the necessary permissions. Which of the following steps should you take to resolve this issue?
Question 3: You are an AWS engineer responsible for optimizing the performance of a DynamoDB table used by a web application. You've identified that the table is experiencing high latency, and you suspect that the DAX cache is not being effectively utilized. Which of the following actions would you take to optimize the cache usage and reduce latency?
Question 4: You are an AWS administrator and you need to ensure that the permissions for the Linux home directory, user's home directory, .ssh directory, and authorized_keys file are correct for an EC2 instance. Which of the following commands should you run to achieve this?
Question 5: Suppose you are an AWS engineer responsible for troubleshooting a connectivity issue between an application and an Amazon RDS database. The application is configured to use a custom parameter group, and the database instance is running in a Multi-AZ deployment. After reviewing the documentation, which of the following steps would you take FIRST to resolve the issue?
Question 6: You are the administrator of an AWS S3 bucket named 'DOC-EXAMPLE-BUCKET'. You have configured CloudFront to serve objects from this bucket. When a user requests the object 'index.html' from CloudFront, they receive an error message saying that the object is not found. Which of the following steps should you take to resolve this issue?
Question 7: You are an AWS administrator responsible for managing access to AWS resources across multiple accounts. You have been tasked with troubleshooting an issue where a user is unable to copy an object from one bucket to another. The error message indicates that the user lacks the necessary permissions. You have identified the following information:
        * The source bucket is owned by Account A.
        * The destination bucket is owned by Account B.
        * The object is owned by Account C.
        * The user attempting to copy the object is in Account D.
    Which of the following steps should you take to resolve the issue?
Question 8: A user has registered a new domain name, but it is not resolving on the internet. They check the domain's status using the whois command and see that it is in 'clientHold' status. What should the user do to make the domain available on the internet again?
Question 9: You are a developer troubleshooting latency issues for an edge-optimized API endpoint in Amazon API Gateway. You have identified the following parts of the connection path and their corresponding durations:
        1. Start of connection to the DNS name resolution: 200 ms
        2. Start of connection to the Transmission Control Protocol (TCP) handshake to connect to CloudFront: 300 ms
        3. Start of connection to the Secure Sockets Layer (SSL) handshake to connect to CloudFront: 400 ms
        4. Start of connection to sending the client HTTP request to CloudFront: 500 ms
        5. Start of connection to the first byte transferred from CloudFront: 600 ms
        6. Total time for the request and response to the API: 1000 ms
        7. Time for API Gateway to process the request and respond to the CloudFront edge location: 300 ms
        8. Time for the integration endpoint to respond to the HTTP request from API Gateway: 400 ms
        9. Time for API Gateway to respond to the CloudFront edge location, and for CloudFront to respond to the client: 300 ms
    Which part of the connection path is the source of the latency for the edge-optimized API endpoint?
Question 10: You are a cloud engineer responsible for deploying a web application using AWS Lambda@Edge and CloudFront. You have associated the Lambda@Edge function with the CloudFront distribution, but you are experiencing 500, 502, and 503 errors. Which of the following steps should you take to troubleshoot the errors?
\end{promptexample}

\vspace{5pt}
\noindent \textbf{Task $t_{\mathit{arx}}$ (Arxiv).} The task~$t_{\mathit{arx}}$ is defined over $13\,000$ ArXiv papers where each paper is represented by its abstract. More precisely, for each of the 13 most commonly used research tags, we randomly sample $1\,000$ papers published before 2021. We require the abstract to have a character length between $,000$ and $1500$. The following example is a question generated for the domain of astrophysics and astronomical instrumentation.

\begin{promptexample}
{
  "question": "The Swift team announced an update to the UltraViolet and Optical Telescope calibration to correct for the loss of sensitivity over time. What is the impact of this update on observations in the three near ultraviolet (UV) filters?",
  "documentation": "A Swift Fix for Nuclear Outbursts. In November 2020, the Swift team announced an update to the UltraViolet and Optical Telescope calibration to correct for the loss of sensitivity over time. This correction affects observations in the three near ultraviolet (UV) filters, by up to 0.3 mag in some cases. As UV photometry is critical to characterizing tidal disruption events (TDEs) and other peculiar nuclear outbursts, we re-computed published Swift data for TDEs and other singular nuclear outbursts with Swift photometry in 2015 or later, as a service to the community. Using archival UV, optical, and infrared photometry we ran host SED fits for each host galaxy. From these, we computed synthetic host magnitudes and host-galaxy properties. We calculated host-subtracted magnitudes for each transient and computed blackbody fits. In addition to the nuclear outbursts, we include the ambiguous transient ATLAS18qqn (AT2018cow), which has been classifed as a potential TDE on an intermediate mass black hole. Finally, with updated bolometric light curves, we recover the relationship of \\citet{hinkle20a}, where more luminous TDEs decay more slowly than less luminous TDEs, with decreased scatter as compared to the original relationship.",
  "choices": [
    "A) It decreases the sensitivity of observations in the UV filters by up to 0.3 mag.",
    "B) It corrects for the loss of sensitivity over time, but the magnitude of the correction varies depending on the filter.",  
    "C) It has no impact on observations in the UV filters.",
    "D) It increases the sensitivity of observations in the UV filters by up to 0.3 mag."
  ],
  "correct_answer": "B) It corrects for the loss of sensitivity over time, but the magnitude of the correction varies depending on the filter."
}
\end{promptexample}

Below, we present a sample of $10$ questions from the ArXiv exam:

\begin{promptexample}
Question 1: Consider the following model for market inefficiency:
    $$\frac{dX_t}{dt} = \mu + \sigma X_t \sqrt{1 - \frac{1}{2} \sigma'^2} \epsilon_t$$
    where $X_t$ is the log market price, $\mu$ is the drift term, $\sigma$ is the volatility term, $\sigma'$ is the volatility of the reasonable price, and $\epsilon_t$ is a standard Brownian motion.
    Which of the following statements is true about the behavior of the market price in this model?
Question 2: A distributed energy resource management system (DERMS) is using network topology identification (TI) to organize and operate widespread distributed energy resources (DERs). The TI function relies only on the measurements available to DERMS. Which of the following approaches is used in the proposed method to improve the resiliency of TI against interruption of communication channels?
Question 3: Consider a network with a limited number of target nodes that need to be controlled. The state variables of the system are associated with the nodes of the network. The goal is to control the target variables as time functions. Which of the following approaches would be most appropriate to achieve this goal?
Question 4: Which of the following statements best describes the virial equation of state of low-density nuclear matter, as presented in the reference?
Question 5: Which of the following best describes the main idea behind the method presented in the manuscript for reconstructing the 3D structure of chromosomes from Hi-C and GAM data?
Question 6: Suppose you are analyzing a large dataset of financial variables and want to test for group-specific heterogeneity in a high-dimensional factor model. Which of the following tests would you use, and why?
Question 7: The unidirectional spin heat conveyer effect in a 200nm thin Yttrium Iron Garnet film is investigated using lock-in thermography. Which of the following statements is true regarding the observed temperature profiles?
Question 8: In the context of the evolutionary prisoner's dilemma game, what can be inferred about the impact of deceitful behavior on the evolutionary outcomes in structured populations?
Question 9: The ATLAS detector at the CERN Large Hadron Collider was used to measure the cross-section of high transverse momentum $W$ and $Z$ bosons produced in $pp$ collisions and decaying to all-hadronic final states. Which of the following statements best describes the method used to reconstruct the boosted $W$ or $Z$ bosons in the analysis?
Question 10: Consider a network of nodes where each node represents a cluster in a competitive environment. The connections between nodes represent non-local interactions between clusters. The probability of a connection between two nodes is $p$. The goal is to devise a survival strategy for an arbitrarily chosen cluster. Which of the following strategies is most likely to be effective?
\end{promptexample}

\vspace{5pt}
\noindent \textbf{Task $t_{\mathit{stk}}$ (StackExchange).} The task~$t_{\mathit{stk}}$ is defined over 977 StackExchange question and answer pairs, randomly sampled from $18$ main tags. To generate a self contained documentation corpus, we artificially join the question and the most upvoted answer, as shown in the documentation example bellow. We only requires the resulting paragraph to have a character length bellow $1500$. The following example is a question generated for the domain of Salesforce development, specifically focusing on Visualforce and formula fields.

\begin{promptexample}
{
  "question": "You are working on a project that requires you to use the `MailingAddress` field from the contact object in your visual force email templates and formula fields. However, you are unable to use this field directly. What should you do instead?",
  "documentation": "### User: I have to use the Contact Object's Mailing address in my code. But I am not able to use it in my visual force email templates or formula fields. Please suggest how can this field be used\\n\\n -\\n\\n### Top Answer: The `MailingAddress` is a special field on the contact - it's a group of multiple fields. Instead you need to use `MailingStreet`, `MailingCity`, `MailingState`, `MailingPostcode` and `MailingCountry`. All these fields together form the `MailingAddress` field.",
  "choices": [
    "A) Use the `MailingStreet`, `MailingCity`, `MailingState`, `MailingPostcode`, and `MailingCountry` fields separately in your code.",
    "B) Use a third-party library to format the mailing address into a single string.",
    "C) Create a custom object that contains the mailing address information and use that object in your code.",  
    "D) Use a combination of the `MailingStreet`, `MailingCity`, `MailingState`, `MailingPostcode`, and `MailingCountry` fields to create a new field that represents the full mailing address."
  ],
  "correct_answer": "A) Use the `MailingStreet`, `MailingCity`, `MailingState`, `MailingPostcode`, and `MailingCountry` fields separately in your code."
}
\end{promptexample}

Below, we present a sample of $10$ questions from the StackExchange exam:

\begin{promptexample}
Question 1: Which of the following is a good resource for finding open-source projects related to processing primitives such as FFT, convolution, correlation, and matrix mathematics for machine vision?
Question 2: Which of the following commands will produce the output *M*M with proper spacing between the two M's?
Question 3: You are developing an Android app that uses the PullToRefresh library. You have added a `PullToRefreshListView` to your layout, but you notice that the white separators between the items (dividers) have disappeared. What is the most likely solution to this problem?
Question 4: You are developing a Java application that uses JavaFX, and you want to distribute it to other computers. You have tried using the `--module-path` and `--add-modules` options to include the necessary JavaFX libraries, but it only works when you use the full path to the SDK library. You have also tried copying the lib file to the application folder and using the path to it, but it still doesn't work. What should you do to make the application run on other computers?
Question 5: You are working on a project that requires you to use the `MailingAddress` field from the contact object in your visual force email templates and formula fields. However, you are unable to use this field directly. What should you do instead?
Question 6: You are tasked with finding all files in a directory that do not have group write permissions. Which of the following commands would you use to accomplish this?
Question 7: You are given a table with date and maximum temperature data for a certain location over a period of several years. You are asked to find the average maximum temperature for each day of the year over the given period.
    Which of the following SQL queries would correctly answer this question?
Question 8: You are an AWS solutions architect, and you have been tasked with designing an IAM policy that allows users from another AWS account to access a specific resource in your account. You have been given the following requirements:
        * The users in the other account should only be able to access the resource if they have been granted permission to do so by the resource owner.
        * The resource owner should be able to grant permission to the users in the other account using IAM roles.
        * The users in the other account should not be able to access any other resources in your account.
    Which of the following options meets all of the above requirements?
Question 9: Which of the following options is the best approach for handling null values when retrieving the ID of a fragment in an Android application?
Question 10: You are given a controller method that removes a user from a group. The method has a parameter `group_id` that is passed as a route parameter. The method uses the `where` method to find the group with the given ID and then calls `destroy` on the resulting object. However, the method is not working as expected and is returning an error.
    Which of the following is the correct fix for the method?
\end{promptexample}

\vspace{5pt}
\noindent \textbf{Task $t_{\mathit{sec}}$ (SEC Filings).} The task $t_{\mathit{sec}}$ is defined on 188 documents submitted yearly to the U.S. Securities and Exchange Commission (SEC) by 10 publicly traded companies, company insiders, and brokers. Each document is segmented in 20 sections, and we only select the ones on ``Business Overview (Section 1)'', ``Risk Factors (Section 2)'' and ``Management's Discussion and Analysis of Financial Condition (Section 9)'', for a total of $493$ sections. We break down these sections into documentation chunks by creating set of $6$ non-overlapping sentences, with a a minimal character length of $1000$ and a maximal of $1400$. Finally, we modify the introduction of the documentation corpus to add some brief context on the company (e.g., ``\textit{Here is an extract of a SEC filing from \{COMPANY NAME\}:)}'').

\begin{promptexample}
{
  "question": "What is the main reason for the establishment of a valuation allowance against the remainder of AMD's U.S. deferred tax assets, net of U.S. deferred tax liabilities, in the fourth quarter of 2002?",
  "documentation": "Here is an extract of a SEC filing from AMD (ADVANCED MICRO DEVICES):
 The 2002 income tax provision was recorded primarily for taxes due on income generated in certain foreign tax jurisdictions and the establishment of a valuation allowance against the remainder of our U.S. deferred tax assets, net of U.S. deferred tax liabilities in the fourth quarter, due to continuing substantial operating losses in the United States. As of December 26, 2004, we had federal and state net operating loss carryforwards of approximately $930 million and $45 million. We also had foreign loss carryforwards of approximately $88 million. We also had federal and state tax credit carryforwards of approximately $246 million and $86 million. The net operating loss and tax credit carryforwards will expire at various dates beginning in 2005 through 2024, if not utilized. We maintain a full valuation allowance against all our net U.S. federal and state deferred tax assets and certain of our foreign deferred tax assets ($694 million at December 26, 2004) because of our history of recent losses.",
  "choices": [
    "A) To record the tax provision for the year 2002.",
    "B) To provide for the utilization of net operating loss carryforwards.",
    "C) To reflect the continuing substantial operating losses in the United States.",  
    "D) To offset the taxes due on income generated in certain foreign tax jurisdictions."
  ],
  "correct_answer": "C) To reflect the continuing substantial operating losses in the United States."
}
\end{promptexample}

\newpage
Below, we present a sample of 10 questions from the SEC Filings exam:

\begin{promptexample}
Question 1: Which of the following is NOT a factor contributing to the AAR CORP's expected increase in revenue in fiscal 2016?
Question 2: What is the likelihood of the pending cases having a material adverse impact on the CECO Environmental Corp's results of operations, liquidity, or financial condition?
Question 3: What is the primary reason for the increase in the net financial instrument position of Air Products and Chemicals between September 30, 2008, and September 30, 2009?
Question 4: According to the SEC filing from Air Products and Chemicals, what is the vesting period for market-based deferred stock units?
Question 5: What is the primary energy source used by Air Products and Chemicals in the production of atmospheric gases such as oxygen, nitrogen, and argon?
Question 6: How does CECO Environmental Corp measure the cost of employee services received in exchange for an award of equity instruments?
Question 7: What is the measurement date for the projected benefit obligations of AAR CORP's pension plans?
Question 8: What was the total expense recorded by Air Products and Chemicals for business restructuring and cost reduction plans in 2012?
Question 9: According to the SEC filing from AMD, how are price reductions handled in relation to product cost?
Question 10: According to the SEC filing, what is the basis for the lease rental rate for the corporate office space leased by Adams Resources & Energy from an affiliated entity?
\end{promptexample}

\clearpage
\newpage

\subsection{Granular Analysis of Results}
\label{sec:granular-results}

\begin{figure}[h]
\centering
\begin{subfigure}{}
    \centering
    \includegraphics[width=0.49\textwidth]{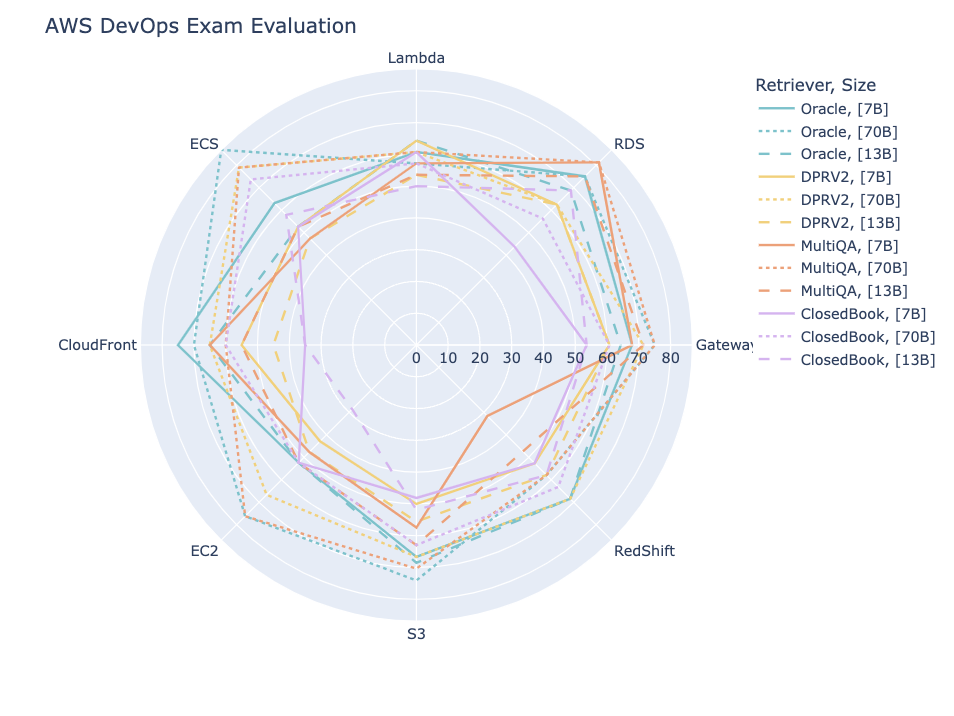}
\end{subfigure}
\hfill
\begin{subfigure}{}
    \centering
    \includegraphics[width=0.49\textwidth]{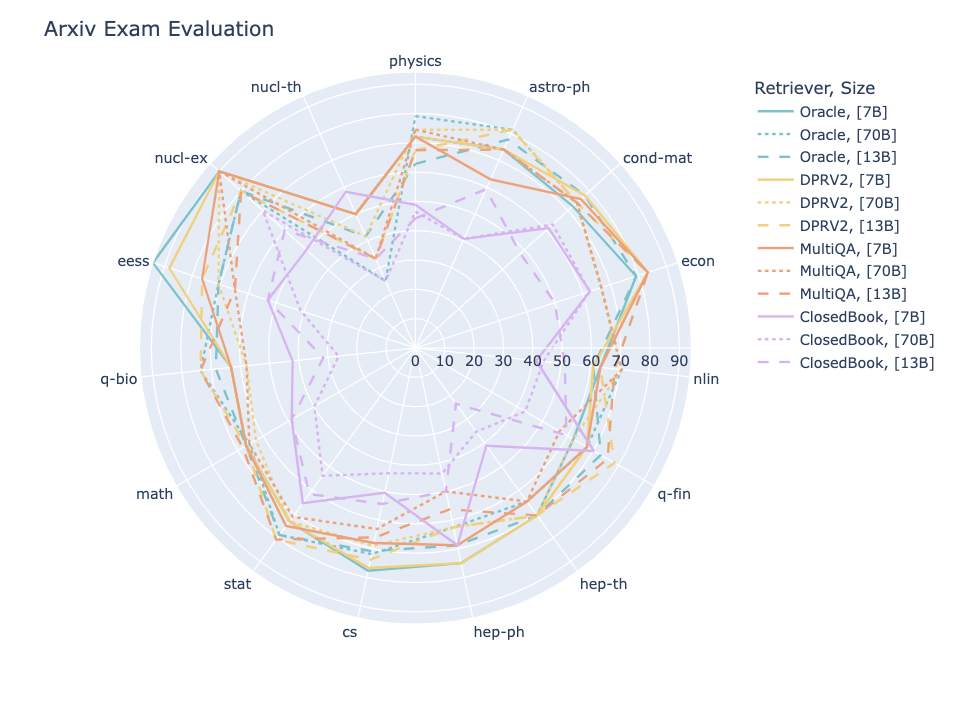}  
\end{subfigure}

\begin{subfigure}{}
    \centering
    \includegraphics[width=0.49\textwidth]{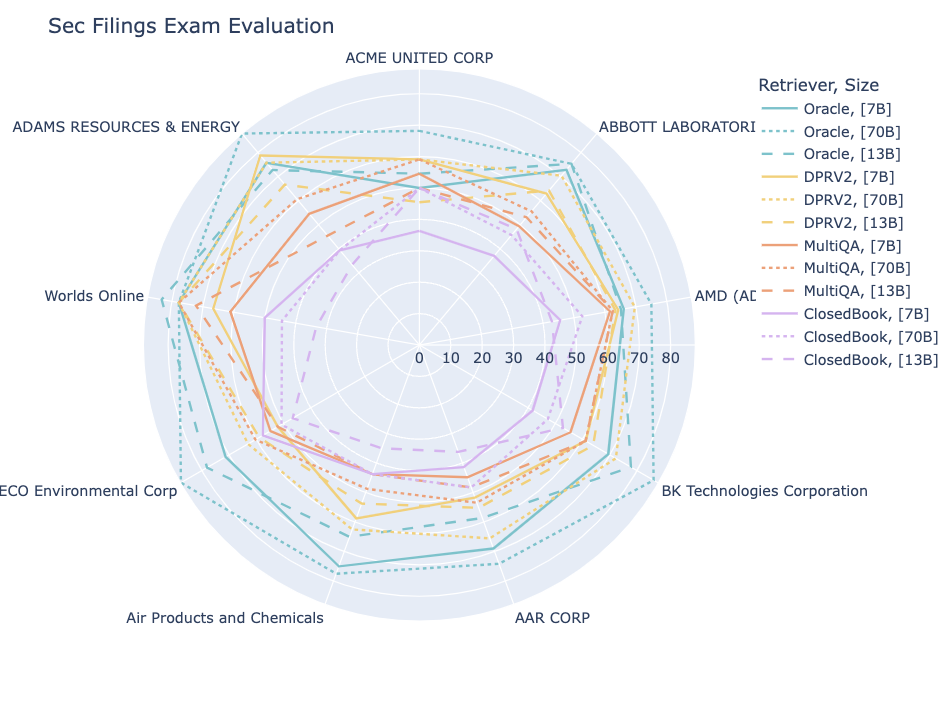}
\end{subfigure}
\caption{Granular results of our exam evaluation for different tasks: $t_{\mathit{ops}}$ on top left, $t_{\mathit{arx}}$ on top right, and $t_{\mathit{sec}}$ on the bottom. Accuracy is reported for different retrieval approaches and retriever sizes, on a \% scale. More details in captions of individual plots. Labels on the diameter shows the different companies. Colors correspond to different retrieval approaches (\texttt{Oracle}, \texttt{DPRV2}, \texttt{MultiQA}, \texttt{ClosedB}, as discussed in Section~\ref{sec:ragpipelines}) and patterns correspond to the base LLM size (Mistral-7B, LlamaV2-13B, and LlamaV2-70B).}
\label{fig:all_radar}
\end{figure}

We next present a granular representation of the exam accuracy for each task (Figure~\ref{fig:all_radar}). Depending on the task and topic, we note that model and retrieval performance greatly varies. These insights are helpful to better understand strengths and limitations of models across several factual directions.

\vspace{15pt}
\newpage
\newpage

\section{Item Response Theory Model Estimation}
\label{sec:hierarchical-irt}

\subsection{Item Response Theory Model Fit}
\label{sec:hierarchical-irt-fit}

To minimize the negative log-likelihood, we leverage L-BFGS-B solver. We initialize at $0$ all the values of $\theta_{m}$ (either for RAG model in the classical IRT or latent variables for the hierarchical model), at $1$ the discrimination $(a_{i})_{i\in\mathcal{Q}}$, at $0$ the difficulty $(b_{i})_{i\in\mathcal{Q}}$ and at $0.25$ the guessing $(c_{i})_{i\in\mathcal{Q}}$.
We enforce the following constraints: $0.1 \leq a_{i} \leq 1.5$, $0.01 \leq b_{i} \leq 1$, $0.2 \leq c_{i} \leq 0.4$ and $-3 \leq \theta_{k} \leq 3$. Finally, as a form of regularization, we experimented with adding a log-normal prior on~$a$, a normal prior on $b$ and a beta prior on $c$, as commonly done in the literature. Yet, these priors led to marginal impact and were removed in the final version. We refer the reader to the source code at \href{https://github.com/amazon-science/auto-rag-eval}{https://github.com/amazon-science/auto-rag-eval} for the detailed implementation.

\begin{table*}[!ht]
\centering
\begin{tabular}{lrrrr}
\toprule
\textbf{Attribute} & $t_{\mathit{ops}}$ (\textbf{DevOps}) & $t_{\mathit{arx}}$ (\textbf{Arxiv}) & $t_{\mathit{stk}}$ (\textbf{StackExchange}) & $t_{\mathit{sec}}$ (\textbf{SEC Filings}) \\
\midrule
Average Exam Accuracy                   & $50.8\% \pm 11.7$ & $60.0\% \pm 12.2$ & $56.1\% \pm 12.2$ &  $53.5\% \pm 11.1$ \\
RMSE - Mean Prediction Baseline       & $0.49 \pm 0.03$   & $0.47 \pm 0.06$   & $0.48 \pm 0.04$ & $0.49 \pm 0.03$ \\
RMSE - IRT                      & $0.44 \pm 0.05$   & $0.42 \pm 0.06$   & $0.43 \pm 0.05$ &   $0.42 \pm 0.04$\\
Discrimination - IRT            & $1.10 \pm 0.33$   & $1.09 \pm 0.37$   & $1.16 \pm 0.44$& $1.01 \pm 0.07$ \\
Difficulty - IRT                & $0.47 \pm 0.39$   & $0.40 \pm 0.44$   & $0.49 \pm 0.46$ & $0.22 \pm 0.17$\\
Guessing - IRT                  & $0.30 \pm 0.01$   & $0.30 \pm 0.09$   & $0.30 \pm 0.09$ & $0.30 \pm 0.09$\\
Theta - IRT                     & $ -0.61 \pm 1.11$ & $0.14 \pm 0.77$   & $-0.10 \pm 1.04$ & $-0.79 \pm 0.87$ \\
\bottomrule
\end{tabular}
\caption{Statistics on Hierarchical IRT Model Fit. We present the average value (plus/minus a standard deviation) either across questions (Discrimination/Difficulty/Guessing), models (Theta) or both (RMSE). As baseline, we compare to a fixed prediction: picking the average exam performance (Mean Prediction Baseline). RMSE refers to Root Mean Square Error.}
\label{tbl:irt_fit}
\end{table*}

Next, Table~\ref{tbl:irt_fit} presents results on the Hierarchical Model fit resulting from the estimation procedure. More precisely, the table highlights:
\begin{itemize}
    \item The average exam scores of all RAG models, including standard deviations (as a column-wise average from Table~\ref{tab:accuracy}), allowing readers to easily assess and compare the difficulty of different tasks.
    \item A comparison of model fit, specifically $P(X=1|\theta)$, between the Hierarchical IRT model and a baseline across various questions, using RMSE. This comparison, showing a consistent improvement in fit for the Hierarchical IRT model, enables readers to gauge the IRT model’s performance on specific tasks.
    \item The average of inferred question-specific parameters from the IRT model, as described in Equation~\ref{eq:irt}, across all exams. This helps elucidate the unique characteristics of individual questions within the broader question corpus. Notably, these values necessitate further normalization for meaningful comparisons across tasks, elucidating, for instance, why the $t_{ops}$ task appears less difficult than $t_{stk}$, despite higher average exam scores.
    \item The average of the inferred model ability $\theta$, across all RAG models for each task. This aggregation aids in understanding how individual model capabilities stack up against the broader model set. Again, direct task comparisons between tasks may not be applicable.
\end{itemize}

 To conclude this section, a noteworthy follow-up question is on the exact relationship between model performance and the length and size of the corpus. More precisely, are models performing better if the size of a corpora to be searched with is smaller/larger? A thorough answer to this question would involve considering the length and chunk size of the corpus as design variables and study the downstream impact on the performance, for a fixed exam. As seen in Table~\ref{tbl:tasks}, we chose 4 tasks with a representative variation in the corpus size (from 977 to 13,000 documents) and document length (from 144 to 254 words). For our initial experiments, we noticed that the variations around these values led to second order differences in performance and thus decided to commit to a given value. For this reasons and given space and focus constraints, we decided to differ an extensive analysis to follow-up work.
 
 Yet, as discussed in Section ~\ref{sec:exam-eval}, our hierarchical framework naturally allows to answer these question in a quantitative way and for any task of interest. More precisely, if we were to address this, we would postulate first a parabolic relation between the ability level variable $\theta$ and the corpus length and assess the model fit. This epitomizes a model where the accuracy first increases with the corpus size, due to documentation overlap and then decreases above a certain size.


\subsection{Task-based Item Response Theory Model Results}
\label{sec:full-irt-plots}

In this section, we present for all four tasks both individual question characteristic curves $p_{i}(\theta)$, item response functions $I(\theta|g_i, d_i, b_i)$ of $p_{i}(\theta)$ and aggregated aggregated Information function $\Bar{I}_{cat}(\theta)$.

\begin{figure}[!h]
\includegraphics[width=\textwidth]{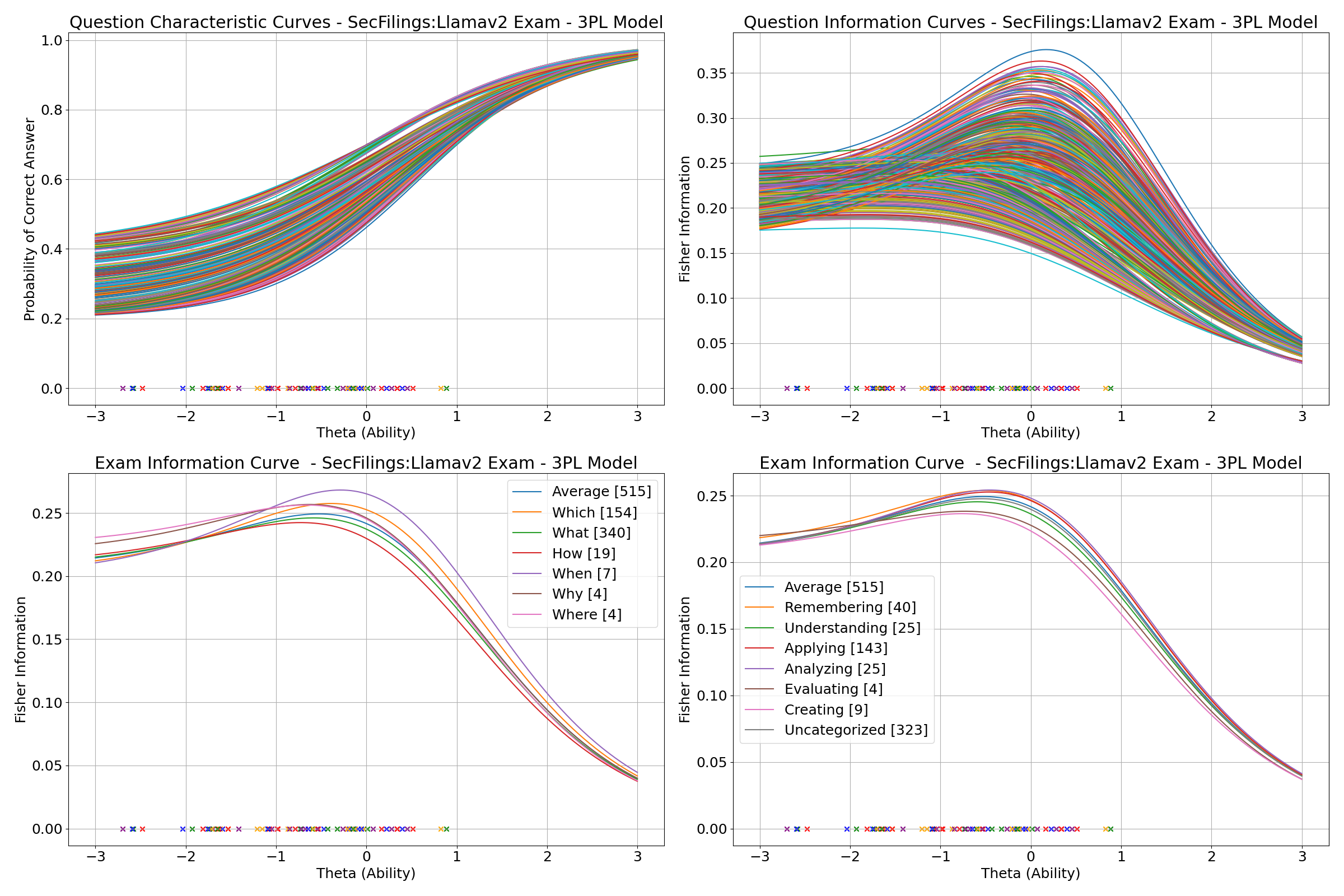}
\caption{Hierarchical IRT Analysis Results for Sec Filings task $t_{sec}$. (\textit{Upper-Left}) Modeling $p_{i}(\theta)$, where each line corresponds to the Item Response Function for a given question $q_{i}\in \mathcal{Q}$. (\textit{Upper-Right}) Aggregated Information function $I(\theta|g_i, d_i, b_i)$ of $p_{i}(\theta)$, where each line to a given question $q_{i}\in \mathcal{Q}$. (\textit{Lower-Left}) Modeling aggregated Information function $\Bar{I}_{cat}(\theta)$, averaged across questions according to semantic taxonomy. (\textit{Lower-Right}) Modeling aggregated Information function $\Bar{I}_{cat}(\theta)$, averaged across questions according to Bloom taxonomy. For all graphs, each cross on the x-axis correspond to a given model ability $\theta_{m}$.} 
\label{fig:irt_sec}
\end{figure}

\begin{figure}[!h]
\includegraphics[width=\textwidth]{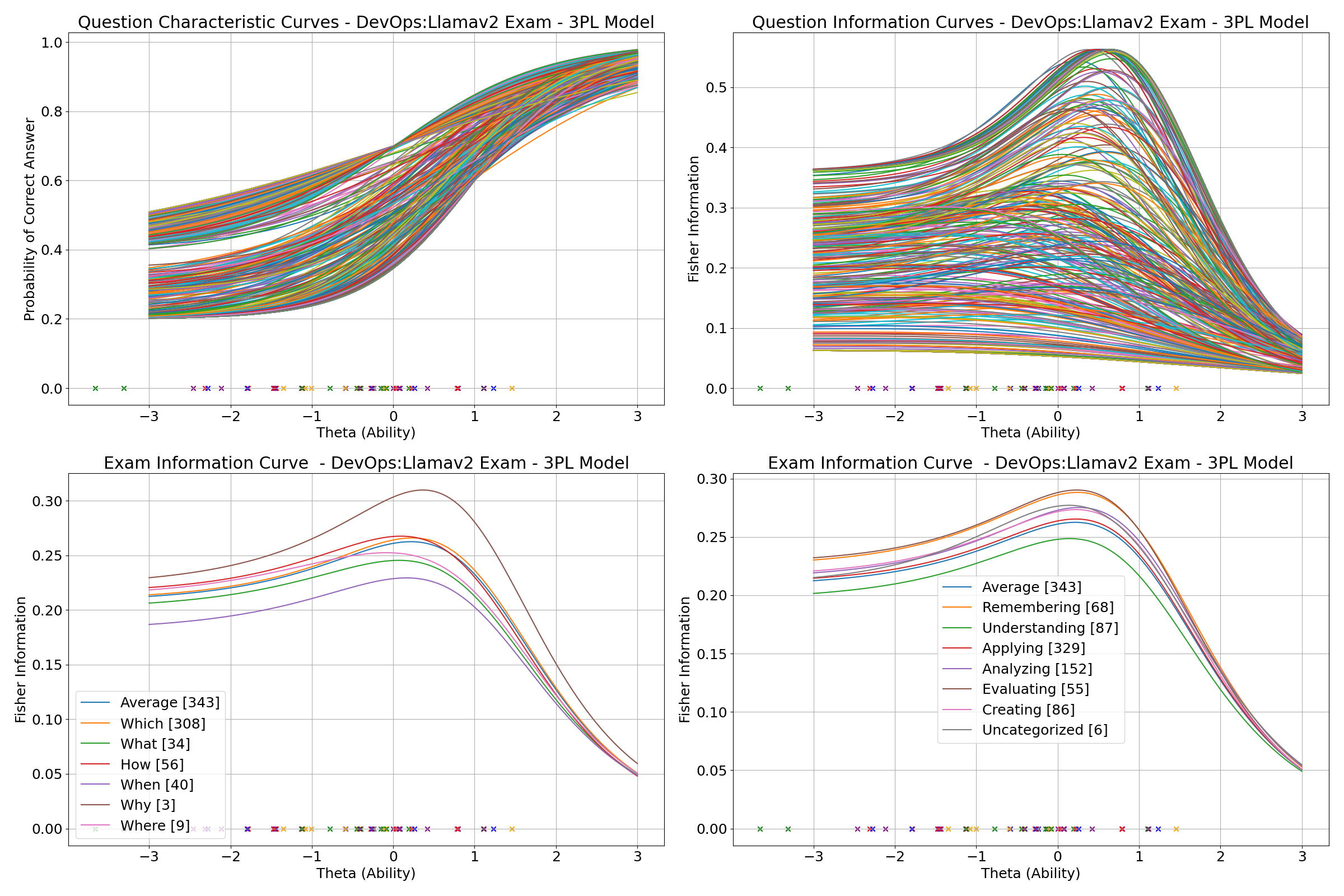}
\caption{Hierarchical IRT Analysis Results for DevOps task $t_{ops}$. (\textit{Upper-Left}) Modeling $p_{i}(\theta)$, where each line corresponds to the Item Response Function for a given question $q_{i}\in \mathcal{Q}$. (\textit{Upper-Right}) Aggregated Information function $I(\theta|g_i, d_i, b_i)$ of $p_{i}(\theta)$, where each line to a given question $q_{i}\in \mathcal{Q}$. (\textit{Lower-Left}) Modeling aggregated Information function $\Bar{I}_{cat}(\theta)$, averaged across questions according to semantic taxonomy. (\textit{Lower-Right}) Modeling aggregated Information function $\Bar{I}_{cat}(\theta)$, averaged across questions according to Bloom taxonomy. For all graphs, each cross on the x-axis correspond to a given model ability $\theta_{m}$.} 
\label{fig:irt_ops}
\end{figure}

\begin{figure}[!h]
\includegraphics[width=\textwidth]{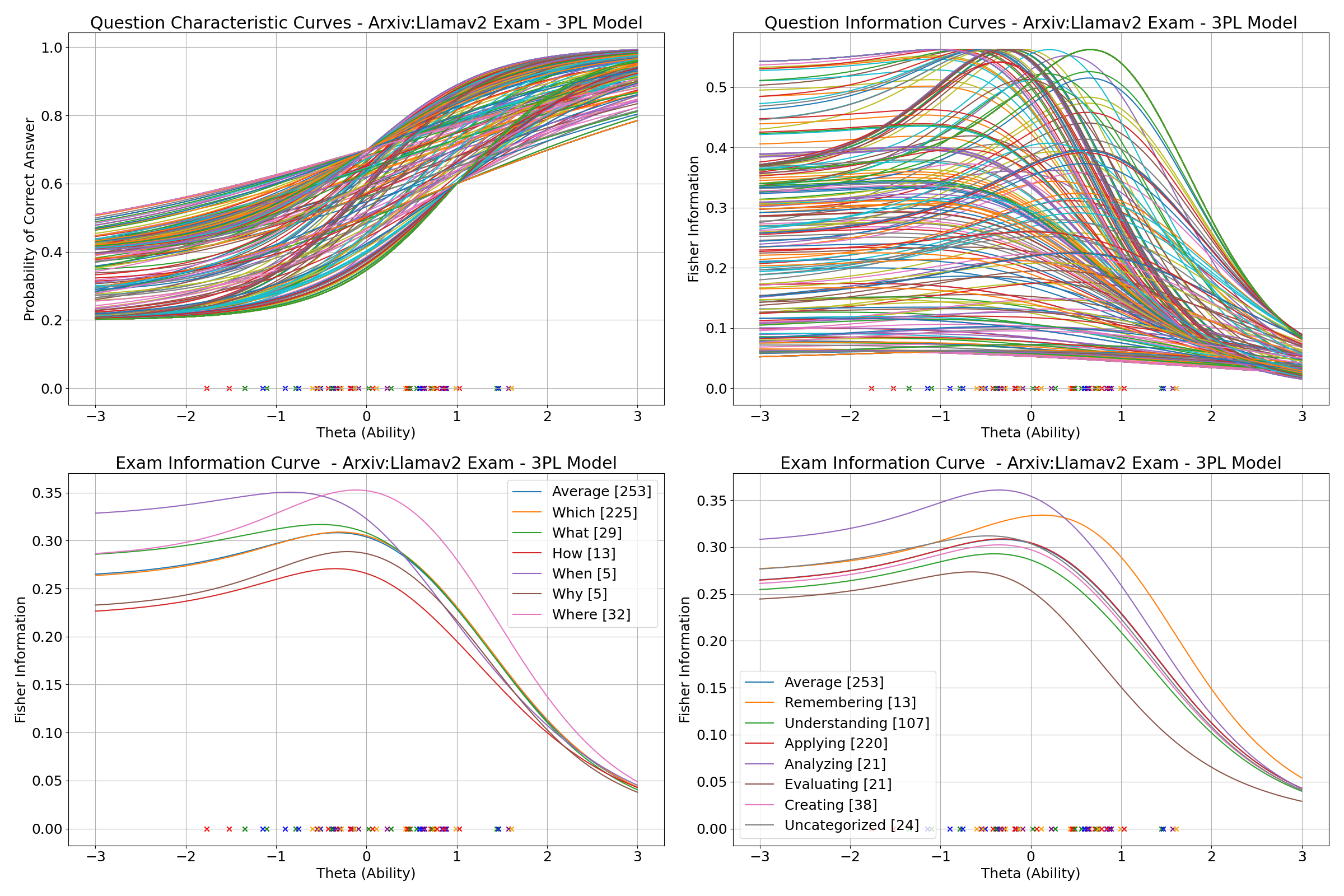}
\caption{Hierarchical IRT Analysis Results for Arxiv task $t_{arx}$. (\textit{Upper-Left}) Modeling $p_{i}(\theta)$, where each line corresponds to the Item Response Function for a given question $q_{i}\in \mathcal{Q}$. (\textit{Upper-Right}) Aggregated Information function $I(\theta|g_i, d_i, b_i)$ of $p_{i}(\theta)$, where each line to a given question $q_{i}\in \mathcal{Q}$. (\textit{Lower-Left}) Modeling aggregated Information function $\Bar{I}_{cat}(\theta)$, averaged across questions according to semantic taxonomy. (\textit{Lower-Right}) Modeling aggregated Information function $\Bar{I}_{cat}(\theta)$, averaged across questions according to Bloom taxonomy. For all graphs, each cross on the x-axis correspond to a given model ability $\theta_{m}$.} 
\label{fig:irt_arx}
\end{figure}

\begin{figure}[!h]
\includegraphics[width=\textwidth]{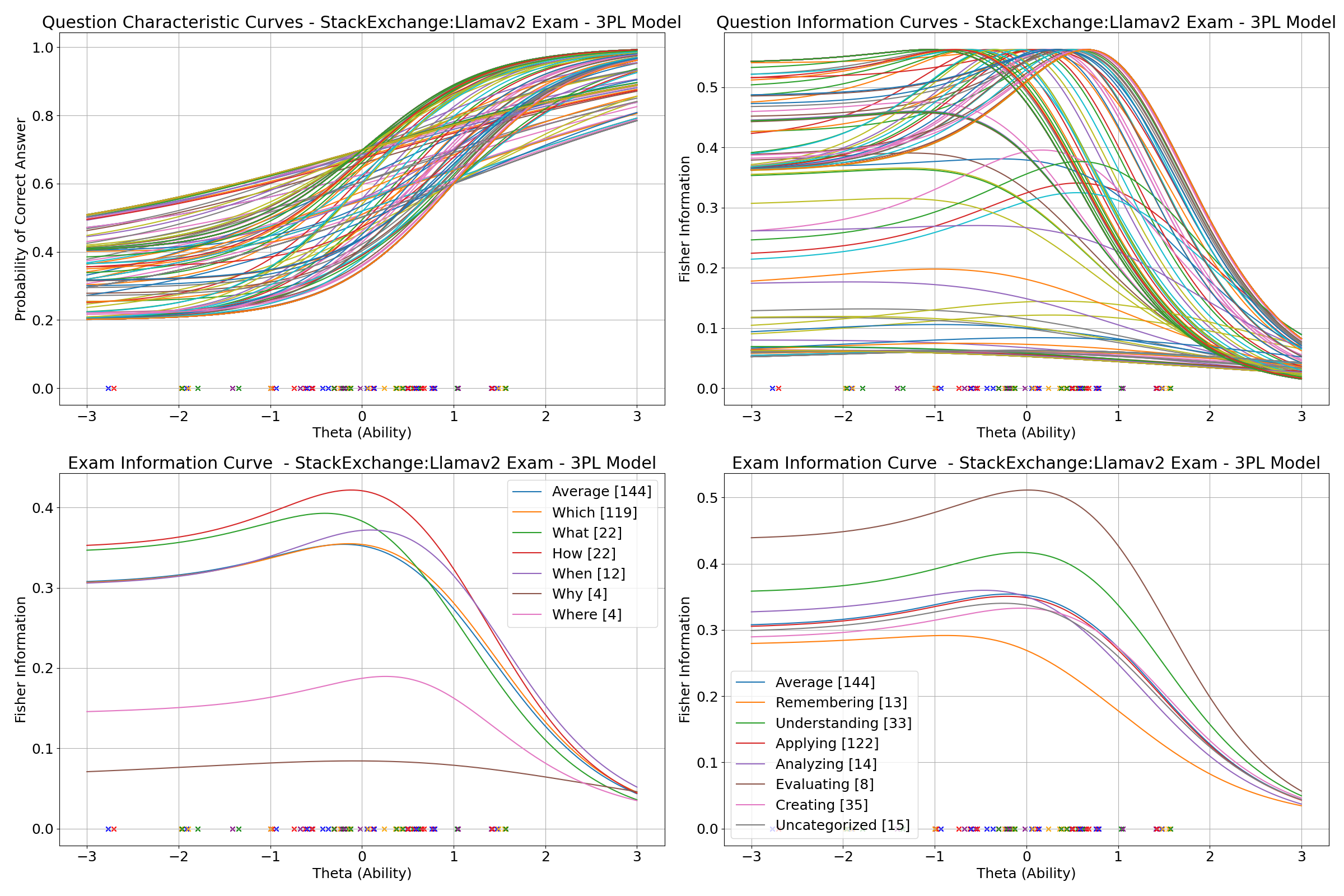}
\caption{Hierarchical IRT Analysis Results for StackExchange task $t_{stk}$. (\textit{Upper-Left}) Modeling $p_{i}(\theta)$, where each line corresponds to the Item Response Function for a given question $q_{i}\in \mathcal{Q}$. (\textit{Upper-Right}) Aggregated Information function $I(\theta|g_i, d_i, b_i)$ of $p_{i}(\theta)$, where each line to a given question $q_{i}\in \mathcal{Q}$. (\textit{Lower-Left}) Modeling aggregated Information function $\Bar{I}_{cat}(\theta)$, averaged across questions according to semantic taxonomy. (\textit{Lower-Right}) Modeling aggregated Information function $\Bar{I}_{cat}(\theta)$, averaged across questions according to Bloom taxonomy. For all graphs, each cross on the x-axis correspond to a given model ability $\theta_{m}$.} 
\label{fig:irt_stk}
\end{figure}

\clearpage
\newpage

\subsection{Iterative Item Response Theory Model}
\label{sec:iterative-irt}

We next present in Algorithm~\ref{alg:iterative-alg} the procedure used to iteratively maximise the informativeness of our exam corpus, by alternatively fitting the IRT model and updating the exam corpus. Here, we simply discard the least discriminative $r^{th}$ quantile of questions (where $r$ is chosen in practice around $10$). Using more sophisticated update technique relying on newer question generation is a fascinating open question and direction for follow-up work.

\begin{algorithm}
\caption{Iterative Exam Improvement with IRT Model}
\label{alg:iterative-alg}
\begin{algorithmic}[1] 
\STATE \textbf{Input:} Initial Exam $\mathcal{Q}_{1}$, Maximal Step Count $k$, Drop Ratio $r$
\STATE \textbf{Initialize:} Fit IRT($\mathcal{Q}_{1}$), infer $(\theta_{m})_{m\in\mathcal{M}}$ and $(g_{i}, d_{i}, b_{i})_{i \in \mathcal{Q}_{1}}$
\FOR{$j = 1$ \textbf{to} $k-1$}
    \STATE \textbf{Exam Update:} Discard the $r^{th}$ first quantile of $(d_{i})_{i\in\mathcal{Q}_{j}}$, the least discriminative questions of $\mathcal{Q}_{j}$ to generate $\mathcal{Q}_{j+1}$
    \STATE \textbf{Model Fiting:} Fit IRT($\mathcal{Q}_{j+1}$), using $(\theta_{m})_{m\in\mathcal{M}}$ and $(g_{i}, d_{i}, b_{i})_{i \in \mathcal{Q}_{j}}$ as initilization.
\ENDFOR
\STATE \textbf{Output:} $\mathcal{Q}_{k}$
\end{algorithmic}
\end{algorithm}

\begin{figure}[!h]
\centering
\begin{subfigure}{}
    \centering
    \includegraphics[width=0.49\textwidth]{img/iterative_exam/Arxiv_fig_llamav2_recursive_irt_3_step4.png}
\end{subfigure}
\hfill
\begin{subfigure}{}
    \centering
    \includegraphics[width=0.49\textwidth]{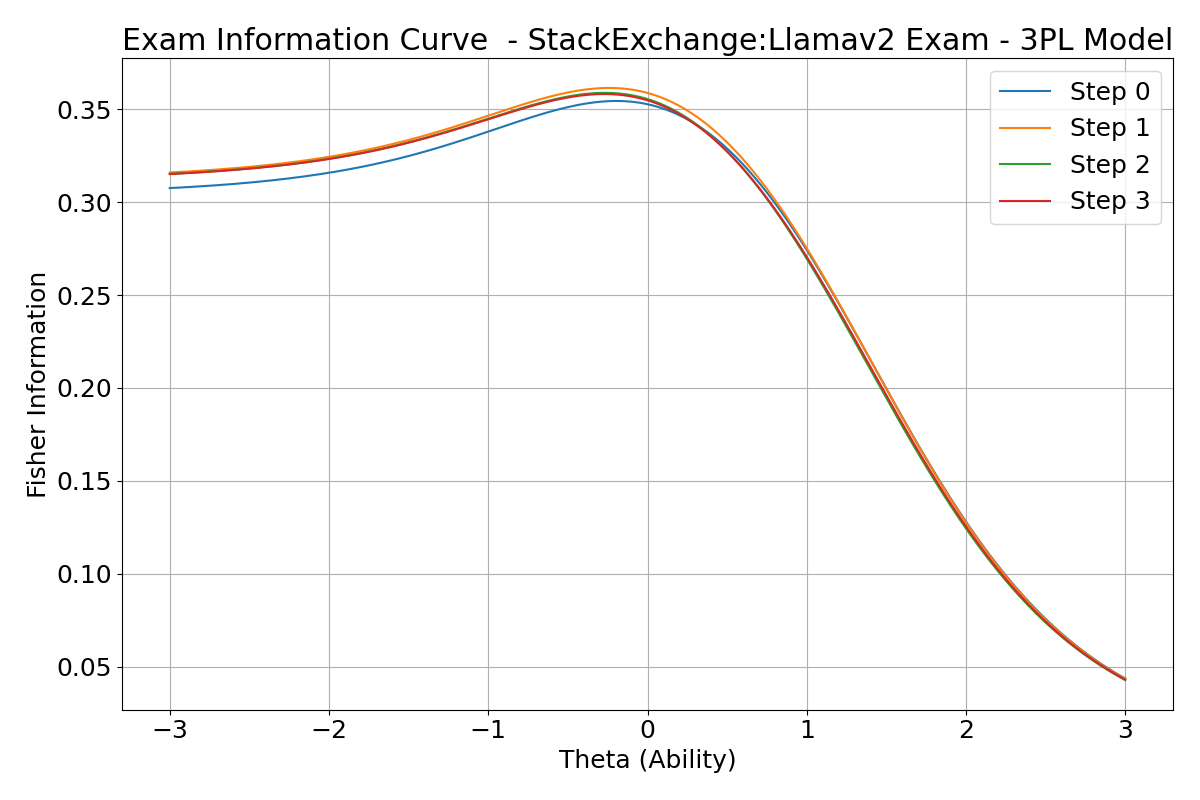}  
\end{subfigure}

\centering
\begin{subfigure}{}
    \centering
    \includegraphics[width=0.49\textwidth]{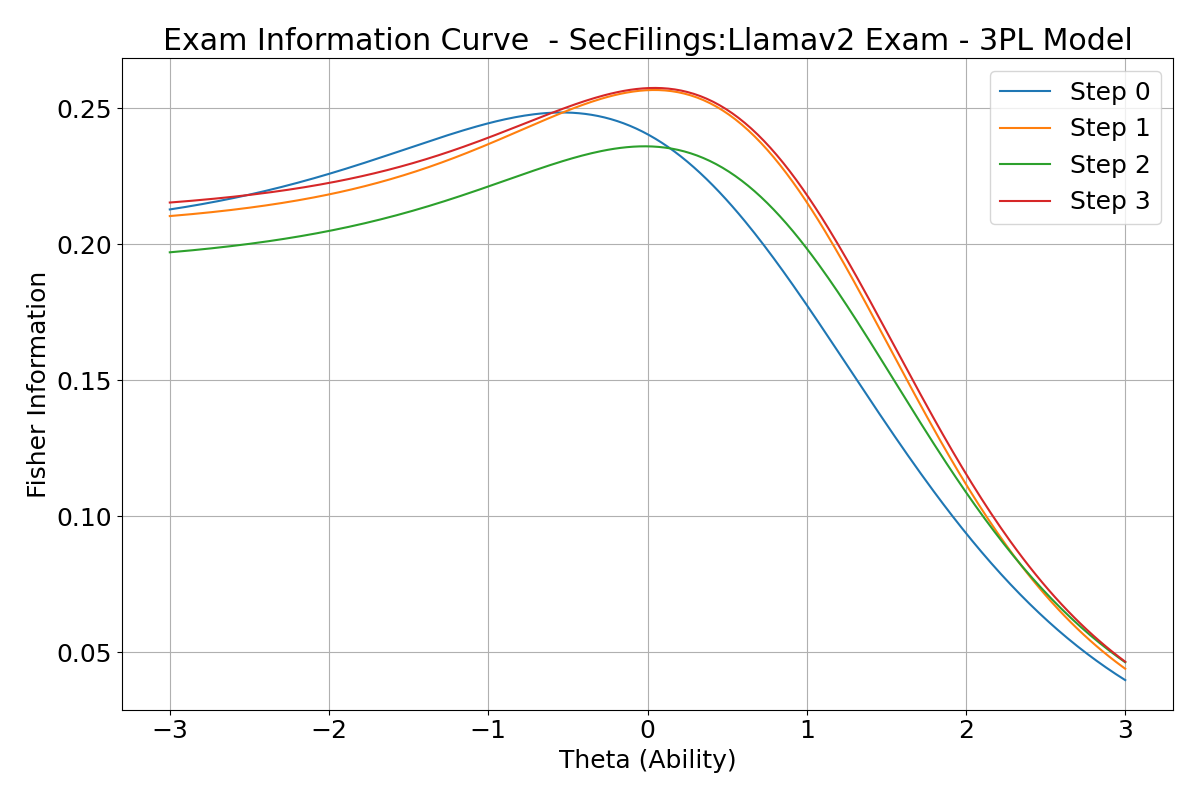}
\end{subfigure}
\hfill
\begin{subfigure}{}
    \centering
    \includegraphics[width=0.49\textwidth]{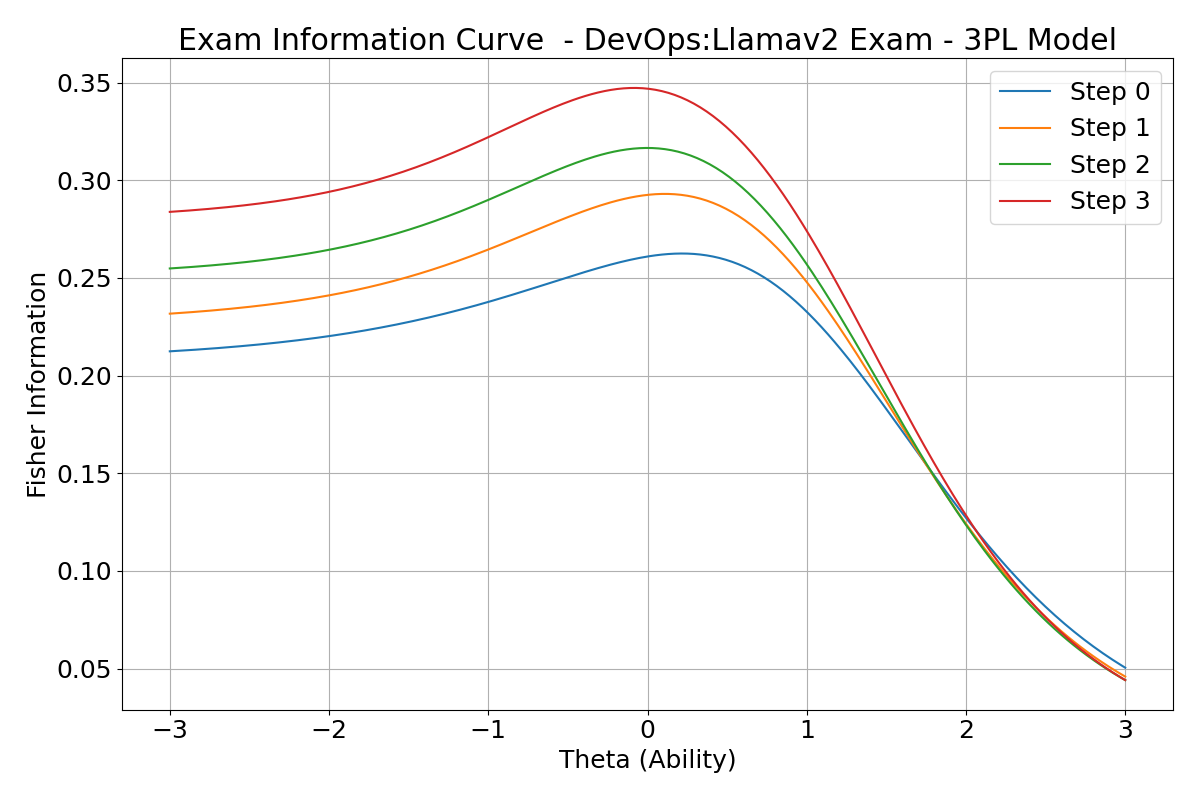}
\end{subfigure}
\caption{Evolution of the Exam aggregated Information function $\Bar{I}_{\mathcal{Q}_{j}}(\theta)$ during the maximization steps $j$. (\textit{Upper-Right}) DevOps Task (\textit{Upper-Left}) StackExchange Task (\textit{Lower-Right}) Sec Filings Task (\textit{Lower-Right}) DevOps Task}
\label{fig:all_iterative_irt}
\end{figure}

Figure~\ref{fig:all_iterative_irt} illustrate the evolution of the exam aggregated Information function $\Bar{I}_{\mathcal{Q}_{j}}(\theta)$ alongside the evolution of the exam corpus. For some tasks such as $t_{arx}$ or $t_{ops}$, we witness a continuous strictly dominating improvement, although mostly in the low to medium ability levels. Such improvement is also witnessed for $t_{stk}$, although convergence happens faster. Finally, for $t_{sec}$, the evolution is non-monotonic and interestingly mostly happens in high ability regions. 

\clearpage
\newpage
\section{Bloom's Revised Taxonomy of Cognitive Domains}
\label{sec:bloom}

Bloom's Revised Taxonomy, an update to the original educational framework by Benjamin Bloom, restructures the classification of cognitive objectives in education. Developed by Lorin Anderson and David Krathwohl, it emphasizes dynamic cognitive processes over static knowledge categories. The revised model, articulated in six progressive levels (Remember, Understand, Apply, Analyze, Evaluate, and Create) moves from basic recall of facts to the sophisticated synthesis of new ideas. This hierarchy guides the design of educational curricula and assessments, focusing on developing higher-order thinking skills and critical analysis.

In this revised framework, the original noun-based categories are replaced with verbs, highlighting active learning processes. For instance, ``Analyze'' involves deconstructing information to understand its components, while ``Create'' represents the peak of cognitive skill, where learners develop new constructs or solutions. This taxonomy is crucial in modern education, providing a foundation for teaching strategies that challenge students' cognitive abilities and prepare them for complex problem-solving across various disciplines.

Table~\ref{tbl:bloom} illustrates the levels of the revised Bloom's Taxonomy, from the lowest to the highest, along with a brief description and examples from the tasks considered of how they might translate into multiple-choice questions.

\begin{table*}[ht]
\small
\centering
\begin{tabular}{lp{4.5cm}p{4.5cm}p{4.5cm}}
\toprule
\textbf{Level} & \textbf{Description} & \textbf{Example Question} & \textbf{Keywords} \\
\midrule
\textbf{Remembering} & Retrieving, recalling, or recognizing knowledge from memory. & What is the reason for AMD being named in the settlement agreement with Philips Semiconductors Corporation? (SecFilings task) & list, identify, name, define, state, mention, recall, label, repeat, recognize \\
\addlinespace
\textbf{Understanding} & Explaining ideas or concepts, understanding the meaning, translating knowledge into new contexts. & Consider a two-dimensional nonlinear Langevin equation with a dissipative, non-potential force that creates a line of stable fixed points (attracting line) touching a line of unstable fixed points (repelling line). In the low-noise limit, the stationary distribution of this system satisfies a large deviation principle with two competing terms. Which of the following statements best describes the nature of these two terms? (Arxiv task) & explain, describe, summarize, predict, interpret, paraphrase, translate, illustrate, rephrase, clarify, check, find, experience, suspect, review, notice, assume, interact, observe, understand \\
\addlinespace
\textbf{Applying} & Using information in new situations or applying knowledge to solve problems. & You are an AWS engineer responsible for scaling a Redis (cluster mode disabled) cluster. You want to reduce costs but ensure minimal downtime for your application. Which of the following scaling actions should you choose, and why? (DevOps task) & demonstrate, apply, use, write, illustrate, solve, show, execute, implement, operate, practice, set, configure, use, try, follow, take, use, run, serve, task, read, operate, work, enable, exist \\
\addlinespace
\textbf{Analyzing} & Breaking information into parts, examining relationships, differentiating between parts. & You are an AWS engineer responsible for deploying and managing an application using Amazon Elastic Container Service (ECS). You have created an ECS service with a task definition, and you want to add tags to the service. However, you are encountering issues with adding tags. Which of the following is the most likely reason for this issue? (DevOps task) & analyze, distinguish, compare, differentiate, examine, test, question, inspect, debate, investigate, manage, differentiate, optimize, troubleshoot, resolve, investigate, compare \\
\addlinespace
\textbf{Evaluating} & Making judgments based on criteria, checking and critiquing. & Which of the following activities contributed the most to AAR CORP's net cash provided by investing activities in fiscal 2015? (SecFilings task) & evaluate, rate, justify, critique, decide, rank, measure, validate, test, assess, evaluate, decide, choose, verify, test, monitor, recommend \\
\addlinespace
\textbf{Creating} & Designing, constructing, planning, producing, or inventing something new based on existing information. & You are developing a Node.js application using the AWS SDK, and you need to use the guard module from the StorefrontControllers package. However, when you try to require the module, you get an error saying that it cannot find the module. What should you do to resolve this issue? (StackExchange task) & design, construct, produce, invent, devise, formulate, originate, assemble, generate, create, design, develop, compose, generate, implement, produce, build, customize, formulate \\
\bottomrule
\end{tabular}
\caption{Bloom's Taxonomy.}
\label{tbl:bloom}
\end{table*}

To cluster the questions in the correct Bloom taxonomy category, we use an algorithm detecting keyword usage, relying on a list extending the one presented in Section~\ref{sec:bloom}. Note that one question can be classified in several categories.


\end{document}